\pdfoutput=1
\documentclass[11pt]{article}
\usepackage{acl}
\usepackage{times}
\usepackage{latexsym}
\usepackage[T1]{fontenc}
\usepackage[utf8]{inputenc}
\usepackage{xurl}
\usepackage{times,latexsym}
\usepackage{url}
\usepackage{amsfonts}
\usepackage{amsmath}
\usepackage{enumitem} 
\usepackage{tabularx}
\usepackage{makecell}
\usepackage{multirow}
\usepackage{graphicx}
\usepackage{subfig}
\usepackage{color}

\usepackage{microtype}
\usepackage[usestackEOL]{stackengine}
\newcommand{\MK}[1]{\textcolor{red}{#1}}

\newcommand{\fmask}{{$F_{\textrm{mask}}$}}
\newcommand{\fbmask}{{$F_{\textrm{bmask}}$}}
\newcommand{\fflip}{{$F_{\textrm{flip}}$}}
\newcommand{\flayer}{{$F_{\textrm{lw}}$}}
\title{Input-specific Attention Subnetworks for Adversarial Detection}
\author{Emil Biju, Anirudh Sriram, Pratyush Kumar, Mitesh M. Khapra \\
        Indian Institute of Technology Madras \\
        \texttt{\{emilbiju@alumni, anirudhs@smail\}.iitm.ac.in }\\
        \texttt{pratyushkpanda@gmail.com}\\ 
        \texttt{miteshk@cse.iitm.ac.in}}
\begin{document}
\maketitle
\begin{abstract}
Self-attention heads are characteristic of Transformer models and have been well studied for interpretability and pruning.
In this work, we demonstrate an altogether different utility of attention heads, namely for adversarial detection. 
Specifically, we propose a method to construct input-specific attention subnetworks (IAS) from which we extract three features to discriminate between authentic and adversarial inputs.
The resultant detector significantly improves (by over 7.5\%) the state-of-the-art adversarial detection accuracy for the BERT encoder on 10 NLU datasets with 11 different adversarial attack types. We also demonstrate that our method (a) is more accurate for larger models which are likely to have more spurious correlations and thus vulnerable to adversarial attack, and (b) performs well even with modest training sets of adversarial examples.
\end{abstract}
\section{Introduction}
Self-attention heads are characteristic of Transformer models. 
Individual attention heads are interpretable in different ways.
One, for a token in an input sentence, we can visualize the attention paid by a head to all other tokens. 
Such attention patterns are attractive linguistically and have come to define roles for attention heads \cite{pande2021heads}.
Two, the output of attention heads from various layers can be probed for their ability to encode information related to the ``NLP pipeline'' \cite{jawahar2019does, tenney-etal-2019-bert, 10.1145/3357384.3358028}.
Three, attention patterns of heads can represent knowledge learnt by a teacher model when distilling to a smaller student model \cite{jiao-etal-2020-tinybert}. 
While individual attention heads are interpretable in the above ways, it is found that attention heads in models such as BERT are over-provisioned and can be pruned. 
For instance, \citet{NEURIPS2019_2c601ad9} showed that a model with 16 attention heads per layer can be pruned to just one. 
\citet{voita2019analyzing} and \citet{budhraja-etal-2020-weak} have shown similar results with different pruning techniques across tasks. 

In the above methods, while interpretation of attention heads is input-specific, pruning of heads is input-agnostic. 
Can these two be combined, i.e., can we prune attention heads in an input-specific manner creating opportunities for interpretation?
We explore this idea to identify an altogether different utility of attention heads - namely \textit{adversarial detection} which is the task of differentiating between authentic and adversarial  inputs.
Specifically, we propose a method to obtain an \textit{input-specific attention subnetwork} (IAS), which is a subnetwork where a subset of attention heads is masked without affecting the output of the model for that input. 
Such subnetworks could vary across inputs representing how the model works for each input.
This is particularly important for adversarial detection, as adversarial inputs do not reveal themselves in \textit{what} the model outputs but may leave tell-tale signs in \textit{how} the model computes this output. 

In this work, we present a technique to efficiently compute IAS and demonstrate its utility in adversarial detection with significantly improved accuracy over all current methods. 
To this end, we propose three sets of features from IAS. 
The first feature, \fmask, is simply the attention mask that identifies if an attention head is retained or pruned in IAS. 
The second feature, \fflip, characterizes the output of a ``mutated'' IAS obtained by toggling the mask used for attention heads in the middle layers of IAS.
The third feature, \flayer, characterizes the outputs of IAS as obtained layer-wise with a separately trained classification head for each layer. 
We train a classifier, called AdvNet, with these features as inputs to predict if an input is adversarial. 

We report results on 10 NLU tasks from the GLUE benchmark (SST2, MRPC, RTE, SNLI, MNLI, QQP, QNLI) and elsewhere (Yelp, AG News, IMDb). 
For each of these tasks, we first create a benchmark of adversarial examples combining 11 attack methodologies like Word order swap \cite{pruthi2019combating}, embedding swap \cite{mrkvsic2016counter}, word deletion \cite{feng-etal-2018-pathologies}, etc.
In total, the benchmark contains 5,686 adversarial examples across tasks and attack types.
To the best of our knowledge, this dataset is the most extensive benchmark available on the considered tasks.
Across all these tasks and attack types, we compare our adversarial detection technique against state-of-the-art methods such as DISP \cite{zhou-etal-2019-learning}, NWS \cite{mozes-etal-2021-frequency}, and FGWS \cite{mozes-etal-2021-frequency}.
Our method establishes the best results in all tasks and attack types, with an average improvement of 7.45\% over the best method for each task. Our detector achieves an accuracy of 80--90\% across tasks suggesting effective defense against adversarial attacks.

Having established the utility of attention heads for adversarial detection, we perform several ablation studies.
First, we compare different combinations of the features demonstrating that they are mutually informative and thus combining them all works best.
Second, we show that CutMix data augmentation \cite{yun2019cutmix} improves accuracy, demonstrating the first use of this method in adversarial detection in NLP tasks.
Third, we show that the detector is more accurate as the size of the language model scales. 
This is encouraging because larger language models are expected to have increased spurious correlations and thus are more vulnerable to adversarial attacks.
Fourth, we show that the detector performs well even for modest training sizes of adversarial examples, suggesting effective generalization. 
In summary, we propose a novel relation between attention heads and adversarial detection. 
The effectiveness of the resultant detector establishes that the mask of attention heads captures critical information about \textit{how} a Transformer model works for a given input.

The rest of the paper is organized as follows. 
We detail our core method of computing IAS in the next section. 
In Section \ref{sec:model_for_adv_det} we discuss the features from IAS for adversarial detection. 
We detail the experimental setup along with the dataset creation process in Section \ref{sec:data_setup}. 
We present our results in Section \ref{sec:results_discussion} and conclude in Section \ref{sec:conclusion}. 

\section{Input-Specific Attention Subnetworks}
In this section, we describe Input-specific Attention Subnetworks (IAS) and the computational approach to identify IAS for a given input. 

\subsection{Notation}
We consider a BERT-style encoder model where each layer consists of multi-headed self-attention and position-wise FFN.
Let an input $x$ consist of $T$ tokens each represented by $d_v$-dimensional vectors.
Let $X_j \in \mathbb{R}^{T \times d_v}$ be the representation at the input of the $j^{th}$ layer.
Let $W_{ji}^Q, W_{ji}^K, W_{ji}^V$ be the projection matrices of the $i^{th}$ self-attention head in the $j^{th}$ layer.
We define $Q_{ji} = X_jW_{ji}^Q, K_{ji} = X_jW_{ji}^K, V_{ji} = X_jW_{ji}^V$ as the query, key, and value corresponding to the head respectively. 
Each self-attention head performs a scaled dot-product attention on the query, key, and value to generate the head's output. The output of all the heads in a layer are concatenated and passed through the FFN.
\vspace{-0.25cm}
\begin{align}\label{multi_head}
    \text{Head}_{ji}(X_j) = \text{softmax}\left(\frac{Q_{ji}K_{ji}^T}{\sqrt{d_k}}\right)V_{ji} \\
\text{Layer}_{j}(X_j) = \text{concat}_i[\text{Head}_{ji}(X_j)]W_j^O
\end{align}
where $d_k$ is the dimensionality of each key vector and $W_j^O$ is a learnable parameter.

A pre-trained model is fine-tuned on a specific task, such as sentiment classification. 
Let $\theta$ be the set of trainable network parameters which are optimized to minimize a task-specific training loss for each input $x$:
\vspace{-0.3cm}
\begin{align}
\mathcal{L}^{\theta}(x)= \mathcal{L}_{CE}(f(x, \theta), y),
\label{eq:loss}
\end{align}
where $f(\cdot)$ is the function computed by the model with parameters $\theta$ for input $x$, $\mathcal{L}_{CE}$ is the standard cross-entropy loss function and $y$ is the expected model output for input $x$. The overall training loss is averaged across all $|x|$ inputs, i.e.,  $\mathcal{L}^{\theta}=\frac{1}{|x|} \sum_{x} \mathcal{L}^{\theta}(x)$.
Let $\widehat{f}(\cdot)$ represent the output class generated from $f(\cdot)$ and $\theta^*$ be the set of optimal network parameters obtained after training.

\subsection{Representing IAS}
\begin{figure}
    \centering
    \includegraphics[scale=0.36]{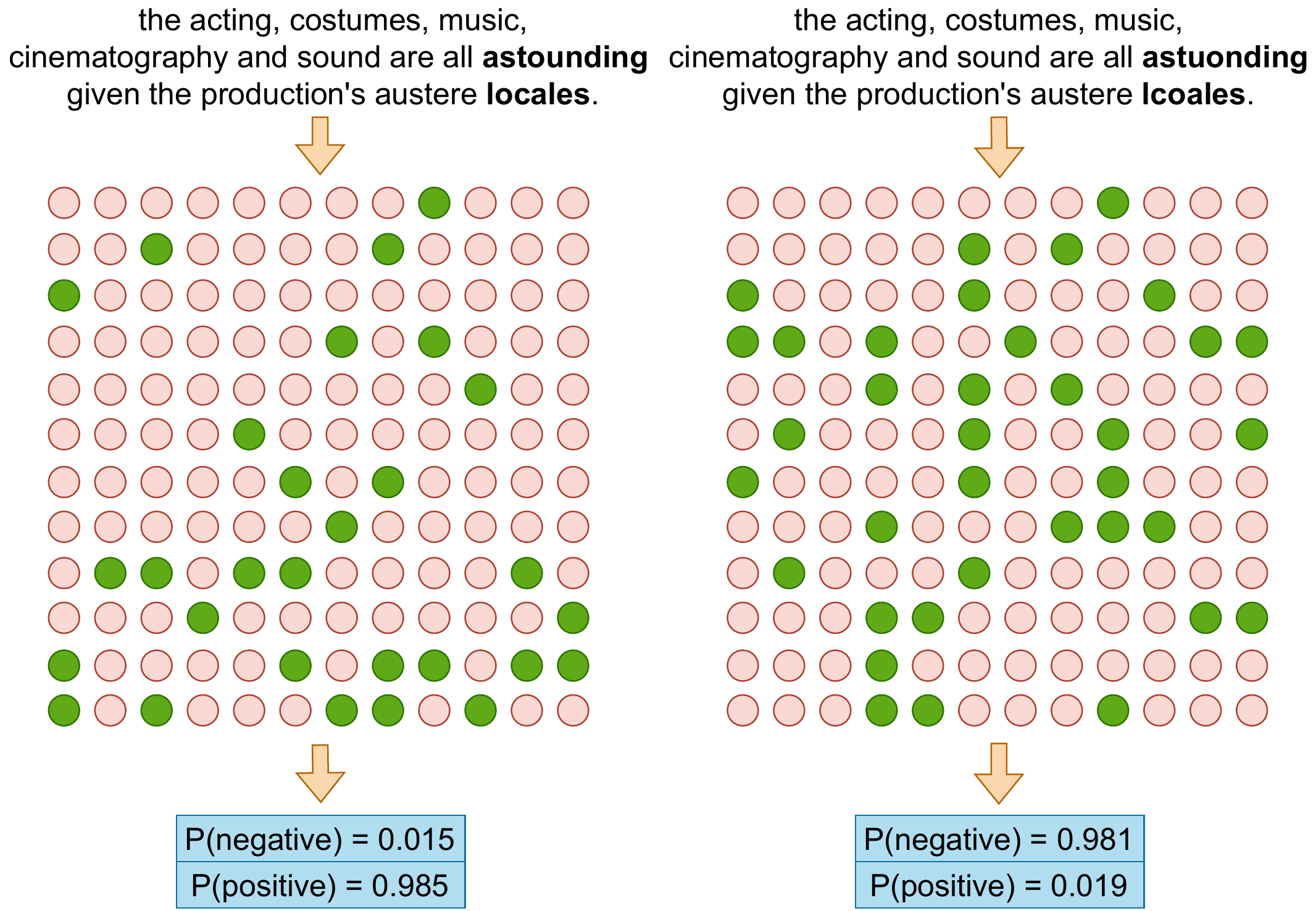}
    \caption{The IAS (with active heads in green) computed for two inputs on the SST-2 task, left is authentic while right is adversarial. Notice how a small adversarial perturbation in the input leads to very distinct subnetworks being computed. The class predicted by each IAS agrees with the prediction of the full network.}
    \label{fig:all_features}
\end{figure}

In an IAS, a subset of attention heads are pruned. 
We represent a continuous relaxation of pruning by modifying Eqn. \ref{multi_head} to weigh the output of each head by a scalar gating value $g_{ji} \in [0,1]$. 
The $j^{th}$ layer of the modified network is given by
\vspace{-0.25cm}
\begin{equation}
\text{Layer}_{j}^{m}(X_j) = \text{concat}_i[g_{ji}\cdot \text{Head}_{ji}(X_j)]W_j^O
\end{equation}
During inference, we constrain the gating values to be binary to characterize either exclusion or inclusion of a head: $g_{ji}$ is replaced by $g^b_{ji} \in \{0,1\}$ which defines the attention mask for the input $x$: $g^b(x) = \{g^b_{ji}\} \in \{0,1\}^{nm}$, where $n$ is the number of layers and $m$ is the number of heads per layer. We represent the output class predicted by the IAS for an input $x$ by $\widehat{f}_g(x, \theta^*, g^b)$.
We call the subset of attention heads that are assigned a gating value of 1 as \textit{active} heads and note that the active heads jointly define a subnetwork, called IAS.
We illustrate IAS with an example.
Figure \ref{fig:all_features} shows the BERT-Base model with 12 layers and 12 heads per layer.
For two specific inputs, the corresponding attention masks are shown with their active heads in green.
Thus, IAS is input-specific and characterizes how the model processes the input in a relatively low-dimensional space of $[0, 1]^{144}$.

\subsection{Computing IAS}
\label{gen_mask}
We compute IAS by treating the gating values as free variables to optimize the task-specific loss (Eqn. \ref{eq:loss}) for a given input $x$.
In this optimization, the network parameters $\theta$ are frozen. 
Each gating value, $g_{ji}$ is defined as $g_{ji} = f_{HC}(p_{ji})$, where $p_{ji}$ is the free variable that is optimized and $f_{HC}$ is a version of the hard concrete distribution \cite{louizos2017learning} given as $\frac{1}{1+e^{\alpha \cdot (log(1-p_{ji})-log(p_{ji}))}}$,
where, $\alpha$=6 gave the best results for our work. 
Let $g$ be the gating vector as optimized by minimizing the loss for a specific input. We need to enforce that $g$ is binary.
Unlike approaches by \citet{voita2019analyzing} and \citet{wang2020interpret}, we do not include a regularization term in the training objective. 
Instead, we retain only those heads for which the gating values ascend the fastest towards 1, as measured after a certain $\eta$ number of epochs. 
Specifically, each binary value $g^b_{ji}$ is derived from $g_{ji}$ after $\eta$ epochs as:
\vspace{-0.3cm}
\begin{align}
g^b_{ji}(x)= 
\begin{cases}
    1,& \text{if } g_{ji}(x)\geq \beta \cdot \max(g(x))\\
    0,              & \text{otherwise}
\end{cases}
\label{eqn:thresh_cases}
\end{align}
where, $\beta (<1)$ is a thresholding parameter and $\max(g(x))$ is the largest among $nm$ gating values. For our work, we set $\eta = 10$ and $\beta = 0.8$.

Two exceptional cases may arise. First, if the binary gating values of all heads in a layer are thresholded to 0, then the largest gating value in that layer is forced to 1 to ensure information flows through the network. Second, if the IAS predicts the wrong class for that input, then $\beta$ is reduced successively in steps of 0.2 until the output of the IAS is correct. For 98\% of the inputs, the subnetwork predicted the target class within $\beta=0.6$.


\section{Model for Adversarial Detection}\label{sec:model_for_adv_det}
In this section, we explain how we extract features from the IAS and the design of the classifier for adversarial detection. We use the term \textit{target class} to refer to the class predicted by the complete fine-tuned network for an input. For authentic inputs, this translates to the true class while for adversarial inputs, this refers to the adversarial class that the model is fooled into predicting. 

\subsection{Attention mask \fmask} \label{sec:model_1}
The IAS identifies a subnetwork through which important information flows for a particular input.
We hypothesize that this flow could be different for authentic and adversarial inputs.
Thus, the first feature we extract, \fmask , is just the pre-activation value $p$ for the gating values of each head in the IAS.
Thus, for a BERT-base model with 12 layers and 12 heads per layer, \fmask~is a 144 dimensional vector.
We also define \fbmask~which uses the binary gated values $g^b$ instead of the real-values.


\subsection{Features from flipping heads in IAS \fflip} \label{sec:mutating_nw}
Adversarial inputs rely heavily on the network architecture and specific parameter combinations to fool the model \cite{wang2019adversarial}. Hence, slight changes to network parameters can render an adversarial perturbation non-adversarial.
We thus hypothesize (and later illustrate in Section \ref{sec:feat_sp_analysis}) that if we flip some of the heads in the IAS, it could significantly change the output for adversarial inputs but not by as much for authentic inputs. 
Which heads should we flip?
We take motivation from studies that show that middle layers of BERT capture syntactic relations \cite{hewitt-manning-2019-structural, goldberg2019assessing} and are multi-skilled \cite{pande2021heads}, making them crucial for prediction.
In contrast, the initial layers are responsible for phrase-level understanding while the last few layers are highly task-specific \cite{jawahar2019does}.
Hence, we choose to flip the gating values $g^b$ of heads in the \textit{middle layers} of IAS, specifically, the middle $\lceil \frac{n}{3} \rceil$ layers, i.e., we drop heads that were earlier active and include earlier inactive heads.
We denote the modified gating vector after flipping as $g^f$.

\vspace{-0.33cm}
\begin{align}
g^f_{ji}= 
\begin{cases}
    g^b_{ji},& \text{if } j \leq \lfloor \frac{n}{3} \rfloor\text{ or } j \geq 2 \lceil \frac{n}{3} \rceil\\
    1-g^b_{ji}, & \text{if } \lceil \frac{n}{3} \rceil \leq j < 2\lceil \frac{n}{3} \rceil\\
\end{cases}
\label{eqn:flipped_IAS}
\end{align}

We run each input $x$ through this mutated subnetwork and obtain a $4$-dimensional feature vector, \fflip~ consisting of the predicted class given by $\widehat{f}_g(x, \theta^*, g^f)$, the target class $y$, the confidence of prediction, and a flag asserting equality between predicted and target classes.

\subsection{Layer-wise auxiliary features \flayer} \label{sec:layerwise_op}
Studies \cite{wang2020interpret, xie2019feature} have shown that intermediate representations of adversarial inputs diverge from those of authentic inputs as we progress into deeper layers. 
This indicates that layer-wise information may be discriminative of adversarial inputs.
Hence, instead of having a single classifier head processing the output of the final layer, we propose to train a classifier head at the output of each layer and use the classes predicted by them as features in adversarial detection.
Specifically, on the fine-tuned complete model, we freeze the standard model parameters to $\theta^*$ and train $n-1$ classifiers separately with a classifier head attached to each of the first $n-1$ layers to predict the target class. Following the convention in Eqn. \ref{eq:loss}, the training loss for the $l^{th}$ classifier head with parameters $\Omega^l$ on input $x$ is given by:
\vspace{-0.3cm}
\begin{align}
\mathcal{L}^{\Omega^l}(x)= \mathcal{L}_{CE}(f^l_g(x, \theta^* \cup \Omega^l, \{1\}^{nm}), y),
\label{eq:loss2}
\end{align}
where $f^l_g(\cdot)$ gives the output class computed by the $l^{th}$ classification head of a network with gating vector $g$. The overall training loss is given by $ \mathcal{L}^{\Omega} = \frac{1}{(n-1)|x|} \sum_x \sum_l \mathcal{L}^{\Omega^l}(x)$. Let $\Omega^*$ be the set of optimal parameters obtained after training.

Then for a given input, we construct the IAS after flipping heads as given by the gating vector $g^f$ and compute the outputs of the $n-1$ layer-wise classifiers, i.e., the output of the $l^{th}$ classifier head is given by $\widehat{f}^l_g(x, \theta^* \cup {\Omega^*}^l, g^f)$.
We then create an $n+1$ dimensional feature, \flayer, which consists of the $n-1$ output labels with two other scalars: (a) the number of these outputs that match the target class, and (b) the number of times these outputs change when traversed in the order of layers.

In summary, we compute the features as follows. 
First, the model is fine-tuned on the task. Then, layer-wise classification heads are trained while keeping the model parameters frozen.
Thus, given an input, we first optimize and compute IAS from which we extract \fmask.
Then, the gating values of the middle layers are flipped and we extract \fflip.
Finally, on the IAS with flipped heads, layer-wise classifier outputs are used to extract \flayer.

\subsection{Classifier for adversarial detection}
\begin{figure}
    \centering
    \includegraphics[scale=0.45]{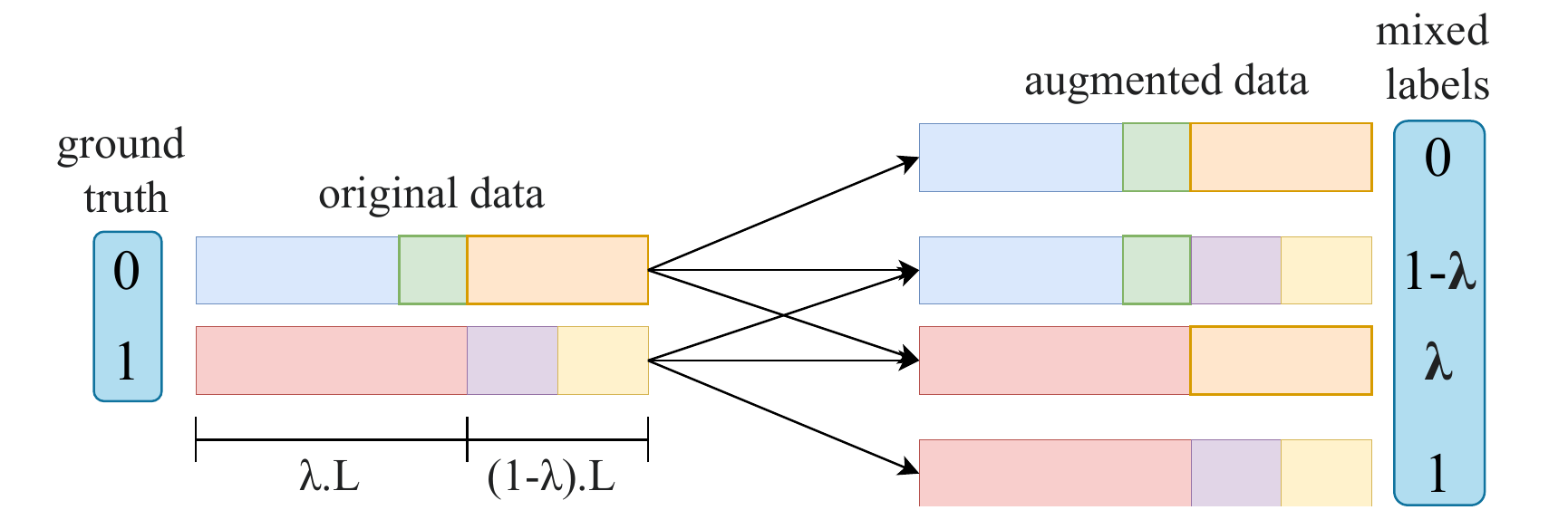}
    \caption{Demonstration of CutMix used to mix patches from two input feature vectors of length L each.}
    \label{fig:cutmix_explain}
\end{figure}

We refer to our classifier as \textit{AdvNet}, which takes as input, an $(nm+n+5)$-dimensional vector $F(x)$ which is the concatenation of \fmask, \fflip, \flayer~and generates a binary output classifying if a given input is authentic or adversarial.
AdvNet consists of two 1-D convolutional layers with ReLU activation, two fully connected layers with sigmoid activation, and a final classification layer with softmax activation. 
Since adversarial inputs are slow and computationally expensive to generate, we employ the CutMix algorithm \cite{yun2019cutmix} for data augmentation. 
In CutMix, we slice out patches from feature vectors of multiple inputs in the training set, each of which could be authentic or adversarial, and combine them to generate new feature vectors. Their respective ground truth labels are mixed in proportion to the length contributed by each patch (see Figure \ref{fig:cutmix_explain}).
Formally, if $\{x_i\}_{i=1}^R$ is a random subset of training set samples, an augmented feature vector from CutMix is defined by $F(\widetilde{x}) = \text{concat}_i [F(x_i)[p_i:p_{i+1}]]$, where $0=p_1<p_2<...<p_{R+1}=nm+n+5$ and the mixed ground truth label is given by $\widetilde{y} = \sum_i y_i (p_{i+1}-p_i)$.
Using soft labels by mixing ground truth labels also offers better generalization and learning speed  \citep{NEURIPS2019_f1748d6b}.


\section{Experimental Setup} \label{sec:data_setup}

\subsection{NLU tasks for evaluation}
\label{sec:data_sec}
We choose the following 10 standard NLU tasks for performing our experimental studies:
SST-2 \cite{socher-etal-2013-recursive},
Yelp polarity \cite{zhangCharacterlevelConvolutionalNetworks2015}, IMDb \cite{maas-EtAl:2011:ACL-HLT2011}, 
AG News \cite{Zhang2015CharacterlevelCN},
MRPC \cite{dolan-brockett-2005-automatically}, 
RTE \cite{wang-etal-2018-glue}, 
MNLI \cite{williams-etal-2018-broad},
SNLI \cite{bowman-etal-2015-large},
QQP\footnote{\url{quoradata.quora.com/First-Quora-Dataset-Release-Question-Pairs}} and 
QNLI \cite{wang-etal-2018-glue, rajpurkar-etal-2016-squad}. We refer the reader to Appendix \ref{appendix:datasets} for further details on these datasets.

\subsection{Dataset creation}
To perform adversarial detection, we require a combined set of authentic and adversarial samples for each task. First, we fine-tune a BERT-based model for each task using its publicly available training set. Then, samples from its test set for which the  fine-tuned model makes correct predictions constitute the set of authentic samples for that task. Second, we generate adversarial samples by attacking the fine-tuned model using a broad set of 11 hard attack types to comprehensively test AdvNet's performance and its generalizability to diverse perturbations.
The attacks include
\textbf{word-level attacks: }deletion \cite{feng-etal-2018-pathologies}, antonyms, synonyms, embeddings \cite{mrkvsic2016counter}, order swap \cite{pruthi2019combating}, PWWS \cite{ren-etal-2019-generating}, TextFooler \cite{DBLP:conf/aaai/JinJZS20} and \textbf{character-level attacks: }substitution, deletion, insertion, order swap \cite{gao2018black}. We use the popular TextAttack framework \citep{morris2020textattack} for implementations of these attacks.
Resulting perturbed samples that successfully fool our complete fine-tuned model constitute the set of adversarial samples for that task. On the combined authentic and adversarial set, we make a 70-10-20 split for creating training, validation and test sets for adversarial detection using AdvNet. 
Our dataset contains a total of 5,686 adversarial inputs across tasks and attack types and is publicly available at \url{https://github.com/emilbiju/Bert-Paths}.

\subsection{Implementation details}

Our adversarial detection model, AdvNet, contains two 1D convolutional layers followed by two fully connected layers. The two convolutional layers have a kernel size of 3 and generate 32 and 16 output feature maps. The two fully connected layers have output dimensions of 32 and 16 with dropout rates of 0.1. We use the binary cross-entropy loss function and the Adam optimizer with a learning rate of 0.001. We train the model for 100 epochs with early stopping on an NVIDIA K80 GPU. 

\begin{figure}[!h]
    \centering
    \includegraphics[scale=0.45]{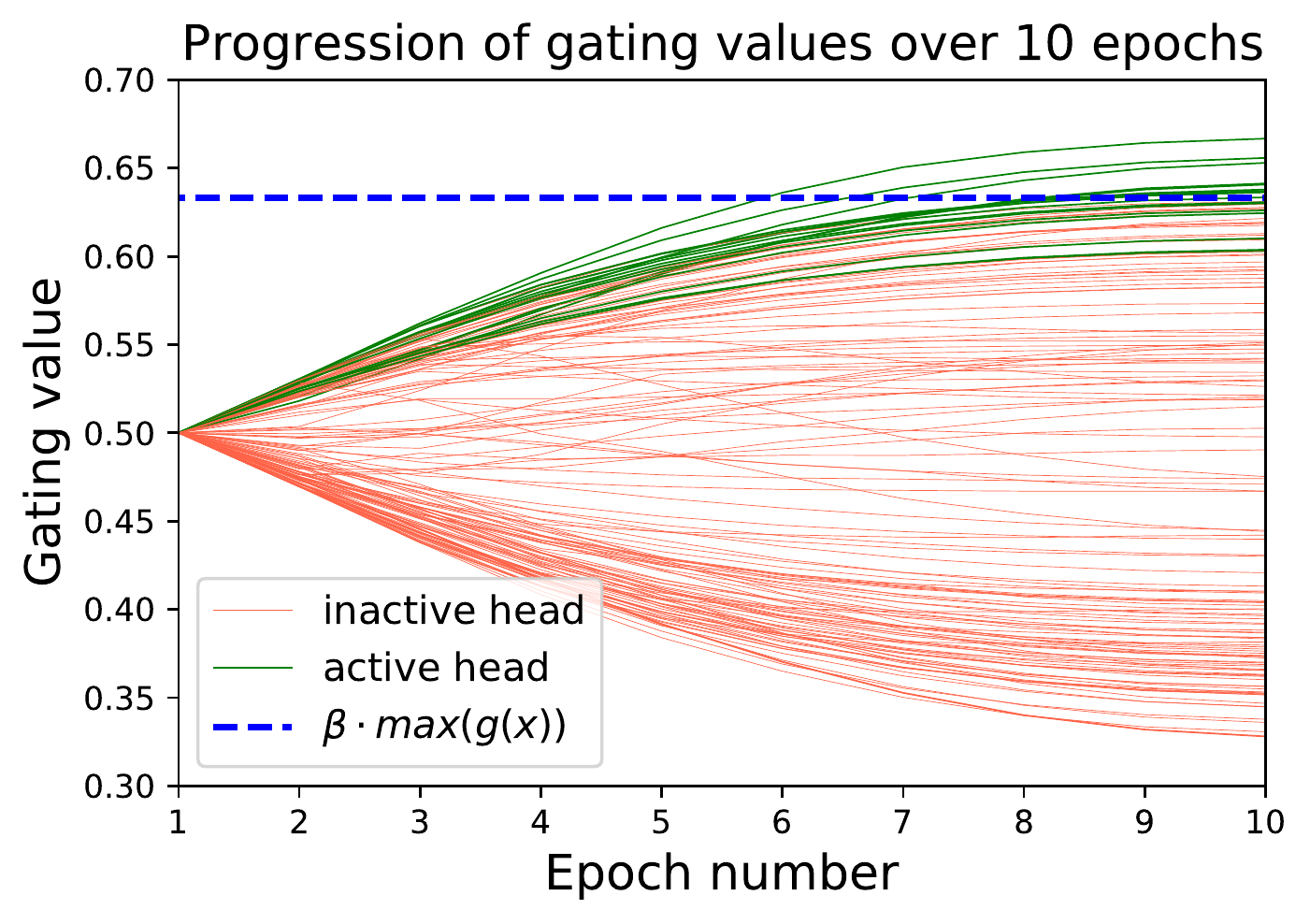}
    \caption{The trajectory of gating values of individual heads during the optimization to compute IAS. Only a few heads (in green) reach the threshold and remain active in IAS.}
    \label{fig:gating_values_vs_epochs}
\end{figure}

\begin{figure}[!h]
    \centering
    \includegraphics[scale=0.45]{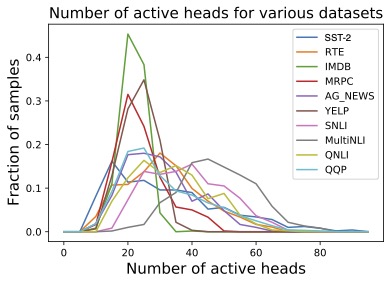}
    \caption{Fraction of inputs with a given number of active heads from BERT-Base. Notice that in most cases, only 20-40 heads out of 144 remain active. }
    \label{fig:gating_values_vs_epochs2}
\end{figure}

\begin{figure*}[!h] 
\centering
    \subfloat{{\includegraphics[scale=0.37]{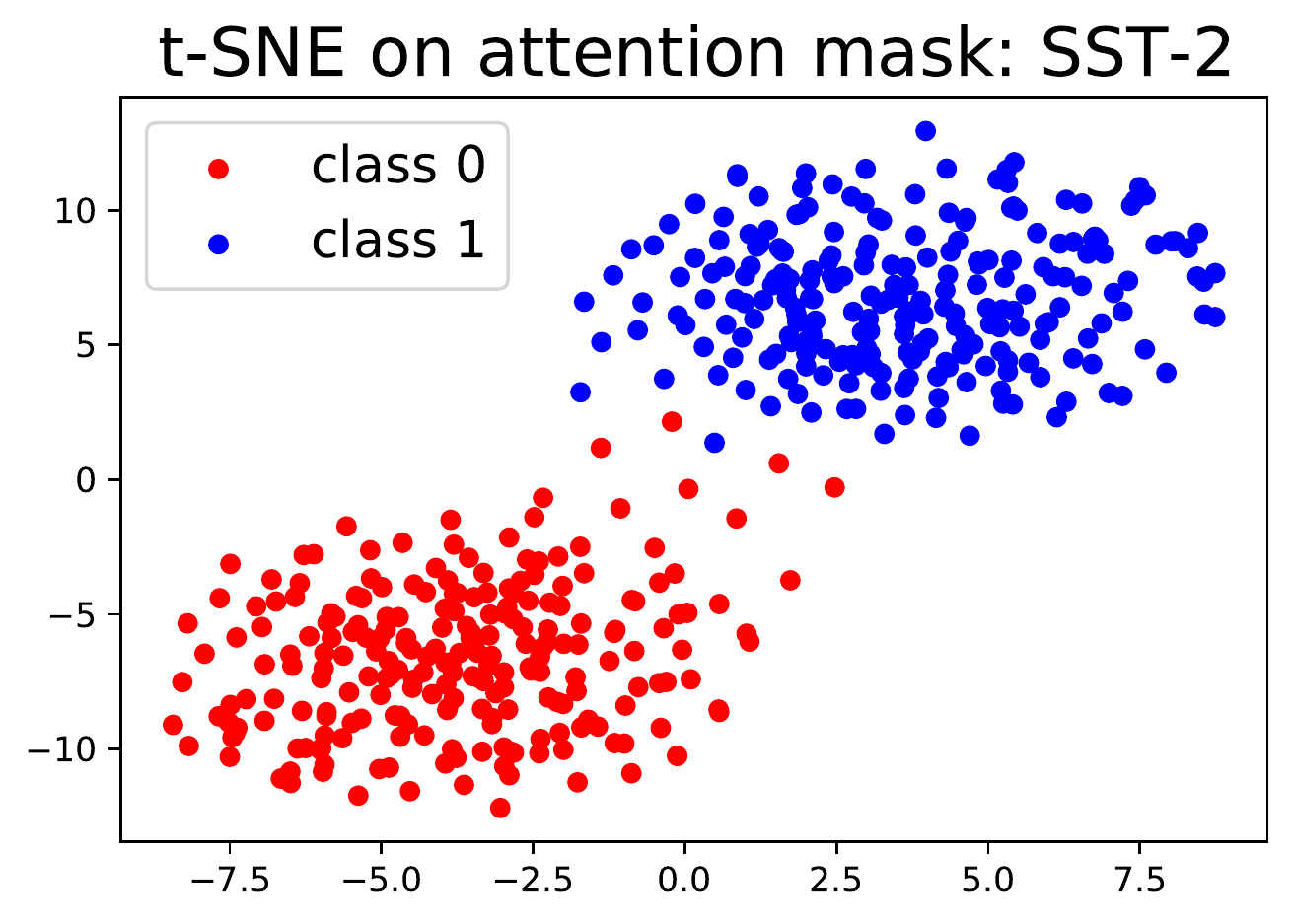}}}
    \quad
    \subfloat{{\includegraphics[scale=0.37]{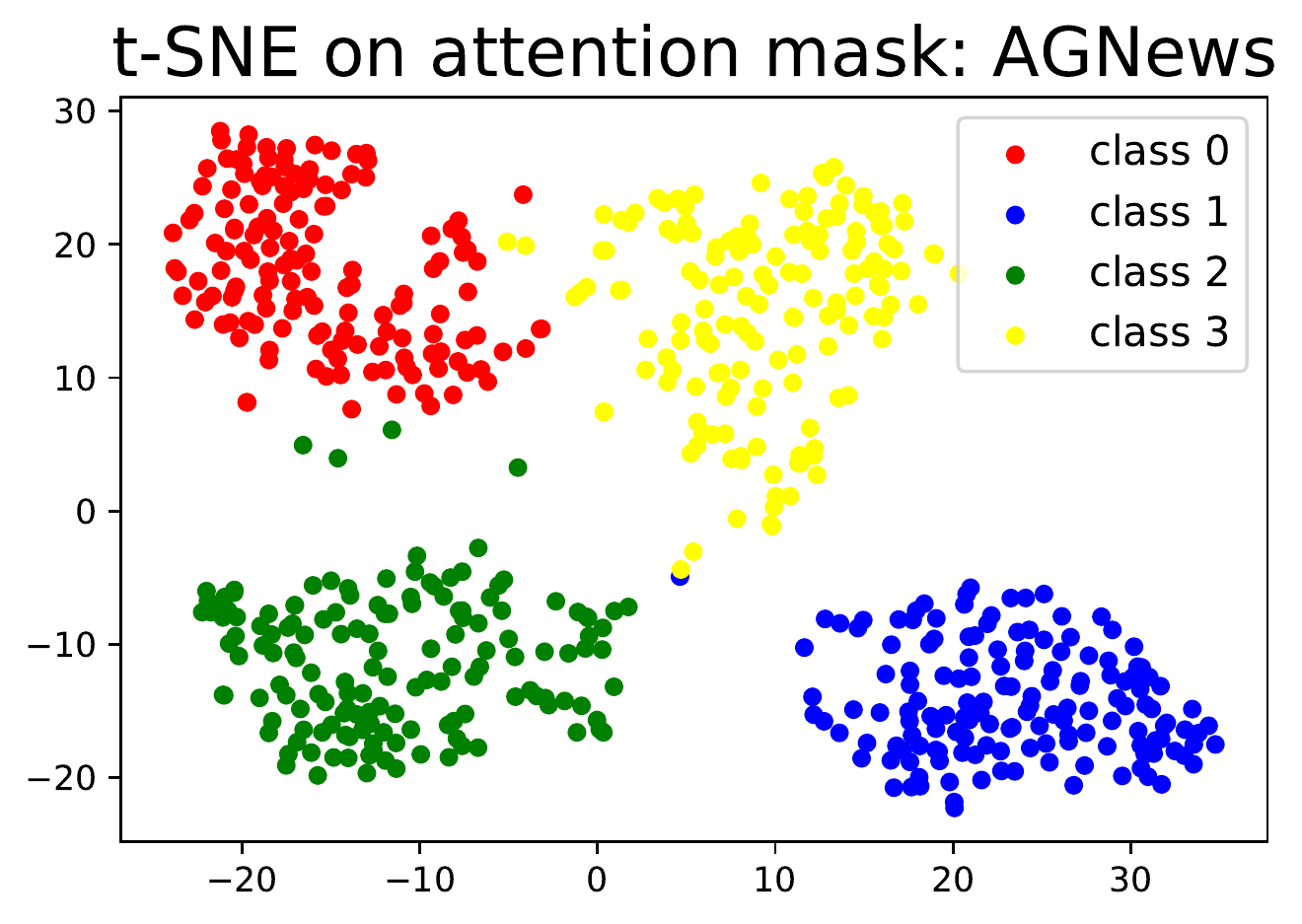}}} 
    \quad
    \subfloat{{\includegraphics[scale=0.37]{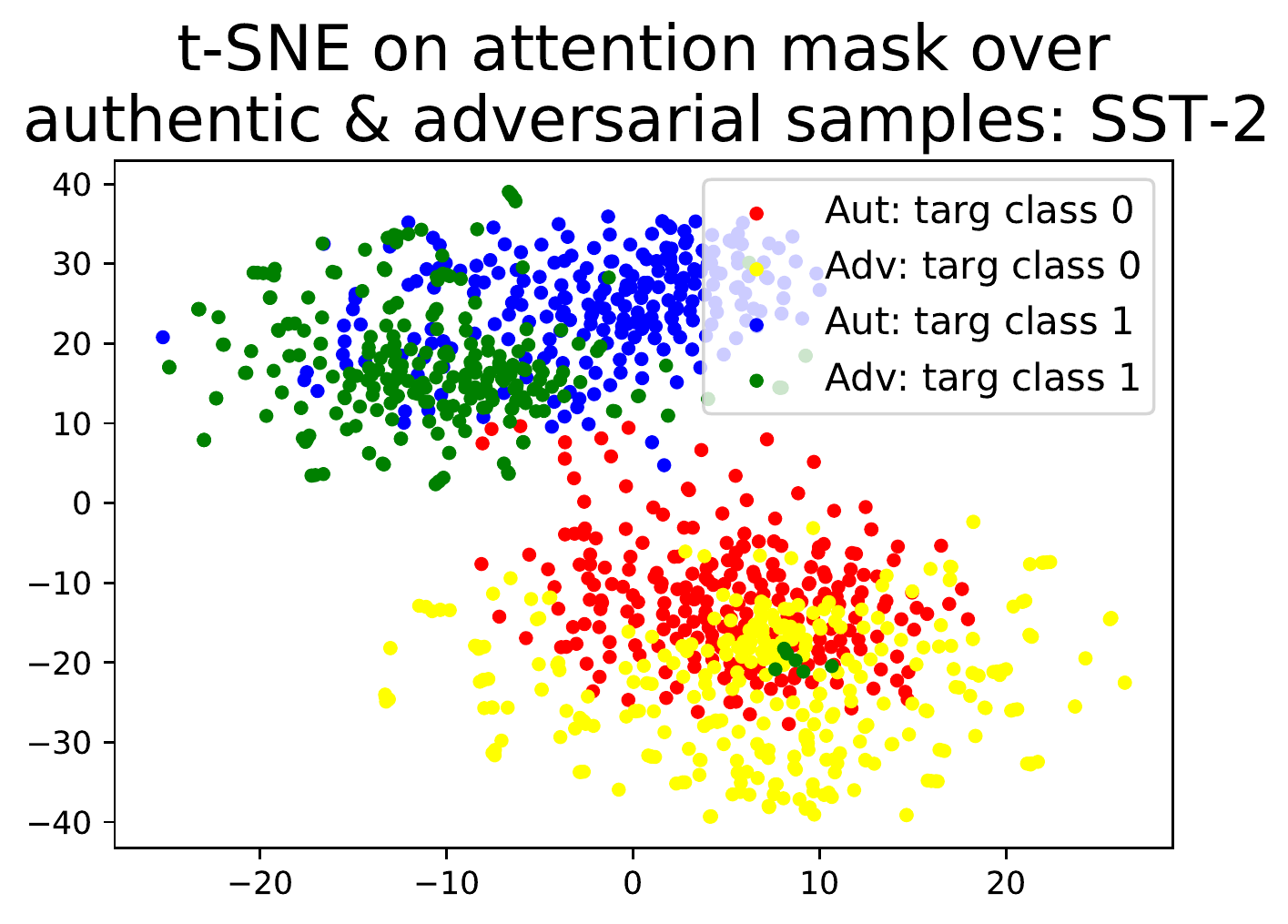}}} 
    \caption{Projections with t-SNE on the attention mask for (a) SST-2, (b) AG News, and (c)  authentic and adversarial inputs. Projection of attention masks are strongly discriminative of class and weakly of adversarial inputs.} 
    \label{fig:tsne_orig}
\end{figure*}

\section{Results \& Discussion} \label{sec:results_discussion}
In this section, we first analyse the IAS (Section \ref{sec:active_heads}) and the constituent features of AdvNet (Section \ref{sec:feat_sp_analysis}). We then perform a comparative study with state-of-the-art adversarial detection methods (Section \ref{sec:adv_det_results}). Lastly, we perform ablation studies to understand the effect of task, model size, feature combinations and training set attacks on the performance of AdvNet (Section \ref{sec:ablation_studies}). Unless otherwise stated, the plots pertain to experiments on the SST-2 dataset with the BERT-Base model.
\subsection{Active heads in IAS} \label{sec:active_heads}
We first check the number of active heads in IAS for a given input. To do so, we plot the progression of gating values with epochs when optimizing them for a given input (see Figure \ref{fig:gating_values_vs_epochs}). We observe that only a small fraction of heads (shown in green) are active at the end of the optimization process, thus resulting in a sparse vector. The green curves that are below the blue (threshold) line correspond to the two exceptional cases discussed at the end of Section \ref{gen_mask}. While the above plot was for a single randomly selected input, in Figure \ref{fig:gating_values_vs_epochs2} we show the fraction of inputs with a given number of active heads for all the datasets used in this work. The relatively small modes and the right skew distributions imply that the extracted IAS are often sparse.

\subsection{Feature-specific analysis}\label{sec:feat_sp_analysis}
We now analyze the individual effectiveness of the three features proposed in Section \ref{sec:model_for_adv_det}. 

\noindent \textbf{Attention mask (\fmask).} We first show that the attention mask is strongly correlated with the input's target class. To do so, we project the binary vector $g(x)$ for each authentic input $x$ onto a 2D-plane using the t-SNE method \cite{JMLR:v9:vandermaaten08a} as shown in Figure \ref{fig:tsne_orig}(a), (b). We observe that inputs from different classes separate into distinctly separate clusters. Thus, the attention mask is discriminative of an input's target class as the choice of active heads depends on it. Interestingly, even if the attention computed for the same word location in two distinct inputs are the same, the heads attending to each word and responsible for generating different output classes are different.

We present a similar plot with both authentic and adversarial inputs in Figure \ref{fig:tsne_orig}(c). We note that adversarial inputs group together with the authentic inputs whose true class is the same as their adversarial/target class. Within clusters of the same target class, there is a only a moderate distinction between adversarial and authentic inputs. But we show in further experiments that a better separation is possible when the complete $nm$-dimensional vector is used as opposed to a 2D projection.

\begin{figure}[!t]
    \centering
    \includegraphics[scale=0.45]{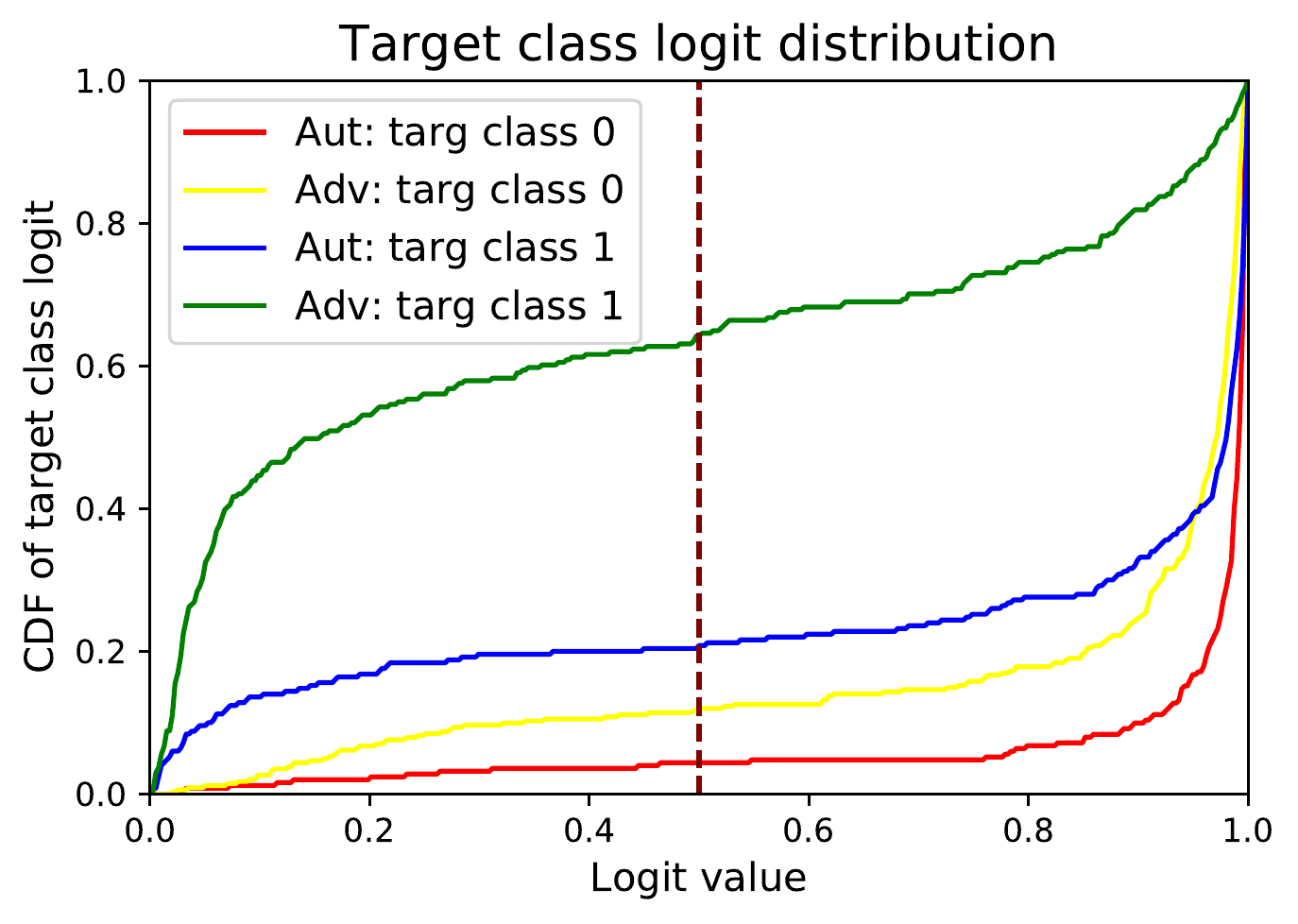}
    \caption{CDF over the target class output logit of the mutated IAS. The large area below the green curve with logit value$<$0.5 corresponds to a large number of adversarial inputs whose mutated IAS predict a non-target class.}
    \label{fig:mut_cdf}
\end{figure}

\noindent \textbf{Features from flipping heads in IAS (\fflip).} For each of the datasets, we compute the percentage of authentic and adversarial inputs which generated non-target class predictions.
We find that the mutated IAS after flipping heads in the middle layers is more likely to predict the correct target class output for an authentic input than an adversarial one.
We also study the confidence of the mutated IAS in making these predictions using a CDF plot (Figure \ref{fig:mut_cdf}) over the output logit  corresponding to the target class.

We observe that \fflip~predicts the target class with higher confidence in case of authentic inputs than adversarial ones.
Specifically, only 9\% of authentic inputs had prediction confidence lower than 0.85 as compared to 20\% of adversarial inputs.
Further, it predicts a non-target class with high confidence for some adversarial inputs. 
For example, 30\% of adversarial inputs with prediction confidence higher than 0.85 gave the wrong prediction.
In contrast, flipping the initial/final layers of the IAS instead of the middle layers did not significantly change the model prediction for either authentic or adversarial samples, making it difficult it to distinguish them.

\begin{figure}
    \centering
    \includegraphics[scale=0.45]{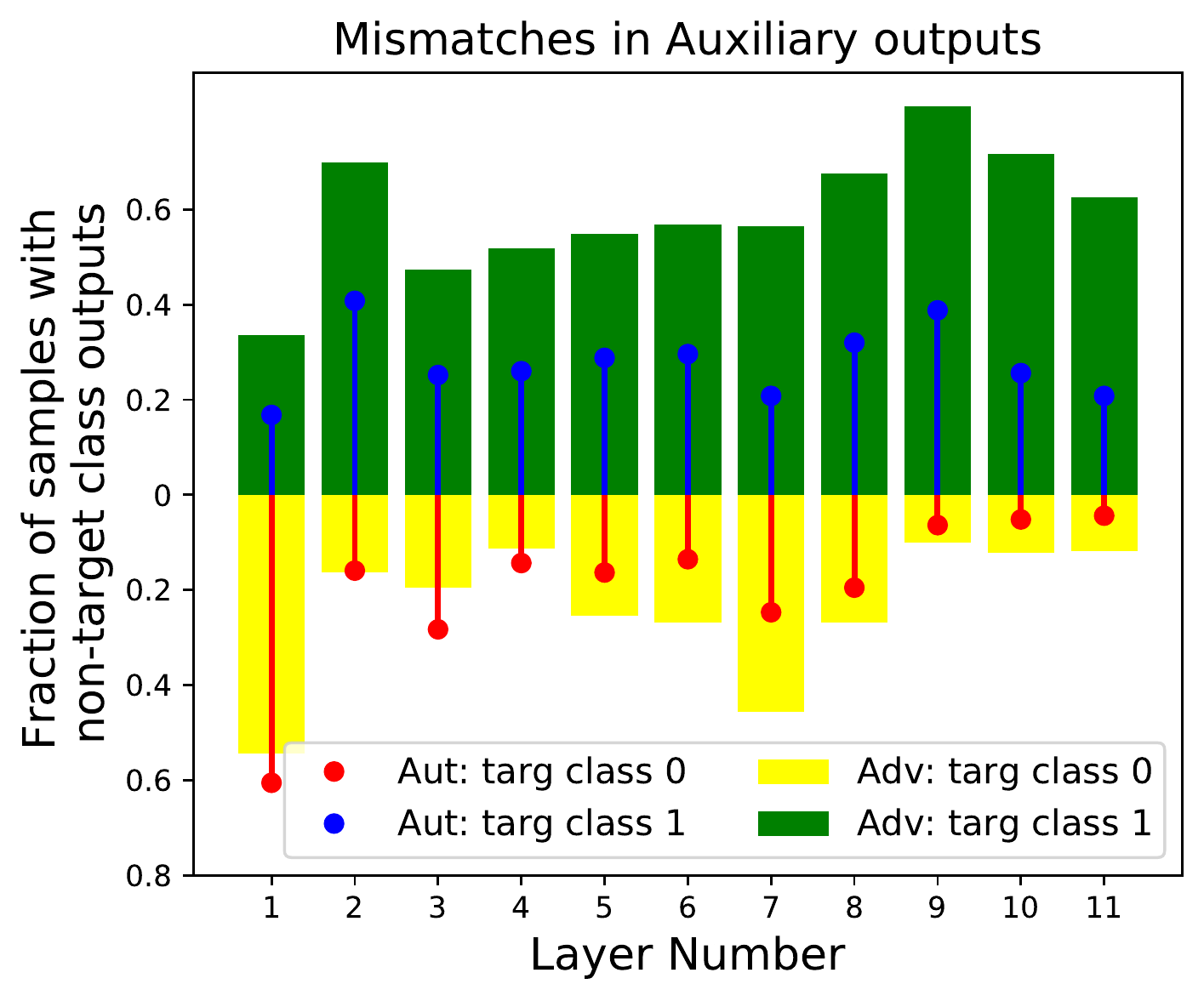}
    \caption{Fractions of authentic and adversarial inputs that generate a non-target class prediction at each layer-wise classification head.}
    \label{fig:aux_outs}
\end{figure}

\noindent \textbf{Layer-wise auxiliary features (\flayer).} In Figure \ref{fig:aux_outs}, we plot the distribution of auxiliary output mismatches (non-target class predictions) across network layers.  
We observe that for most layers, the fraction of authentic inputs having target class predictions is higher than adversarial inputs. 
The differences are particularly large for the last few layers.
On average across datasets, we observed that 52.5\% of adversarial inputs generate more than 2 auxiliary output predictions that do not match the target class while only 23.1\% of authentic inputs do the same.
Additionally, when traversing the layer-wise outputs in order, we observed that the output predictions of adversarial inputs switch among possible classes more often than for authentic inputs (see Appendix \ref{appendix:analysing_fflip}). These observations justify the features that we include in \flayer.

Based on the above analyses, we have demonstrated that all 3 features of IAS are informative for adversarial detection.
Our results in the next section corroborate these findings.

\setlength{\tabcolsep}{4pt}
\begin{table*}[!t]
\centering
\begin{tabular}{c c c c c c c c c c c}
\hline
\textbf{Model}&\textbf{SST-2}&\textbf{Yelp}&\textbf{AG News}&\textbf{MRPC}&\textbf{IMDb}&\textbf{SNLI}&\textbf{RTE} & \textbf{MNLI}&\textbf{QQP}&\textbf{QNLI}\\\hline
{FGWS} & 71.93 & 78.36 & 70.41 & 69.85 & 75.98 & 75.41 & 71.23 & 60.23 & 73.52 & 78.14\\ 
{NWS} & 70.31 & 74.72 & 65.62 & 68.02 & 65.72 & 71.82 & 64.27 & 56.94 & 70.20 & 74.58\\ 
{DISP} & 68.73 & 70.15 & 66.38 & 62.22 & 75.23 & 72.92 & 66.40 & 59.34 & 69.86 & 76.92\\ 
{FreeLB} & 77.60 & 82.54 & 75.55 & 72.41 & 79.85 & 79.80 & 64.29 & 58.10 & 65.69 & 76.40\\  \hline
\textbf{AdvNet} \\
{w/ BERT-Small} & 78.57 & 76.72 & 78.63 & 75.05 & 74.09 & 72.07 & 73.64 & 64.26 & 68.71 & 74.47\\ 
\textbf{w/ BERT-Base} & \textbf{90.74} & \textbf{87.68} & \textbf{91.78} & \textbf{84.61} & \textbf{81.18} & \textbf{82.50} & \textbf{80.43} & \textbf{72.61} & \textbf{75.27} & \textbf{86.07}\\  \hline
\end{tabular}
\caption{Comparison of the adversarial detection accuracy of AdvNet using features extracted from fine-tuned BERT-Small and BERT-Base models with other state-of-the-art approaches for adversarial detection.}
\label{tab:adnet_compare}
\end{table*}
\addtolength{\tabcolsep}{-2pt} 

\subsection{Performance on Adversarial Detection}
\label{sec:adv_det_results}
Following the observations in the previous section, we use AdvNet with the identified features for adversarial detection. We compare the  performance of AdvNet with the current state-of-the-art approaches for detecting adversarial inputs for BERT-based models, \textit{viz.}, \textbf{FGWS} \citep{mozes-etal-2021-frequency}, \textbf{NWS} \citep{mozes-etal-2021-frequency}, \textbf{DISP} \citep{zhou-etal-2019-learning} and \textbf{FreeLB} \citep{Zhu2019FreeLBEA}. We briefly describe these methods in Appendix \ref{appendix:sota_methods}.

As seen in Table \ref{tab:adnet_compare}, AdvNet significantly outperforms existing approaches across all 10 datasets with an average improvement of 7.45\%. 
We report an improvement of 6.53\% for the 3 sentiment analysis datasets (SST-2, Yelp, IMDb), 8.05\% for the 4 NLI datasets (RTE, SNLI, MNLI, QNLI) and 6.98\% for the 2 paraphrase detection datasets (MRPC, QQP) over the respective best methods.

Another baseline that we compare with is \textbf{Certified Robustness Training} \citep{Jia2019CertifiedRT}. While this work is not aimed at adversarial detection, it provides bounds on model robustness for word substitution perturbations. For making a comparison with our work, we note that the fraction of adversarial samples that are correctly detected as adversarial translates to robustness for binary classification tasks. We report robustness of 87\% for word substitution-based attacks and 81\% across all 11 attacks for IMDb, while the best upper bound obtained through certified robustness training is 75\%.

When comparing across datasets, we observe that AdvNet performs better on simpler sentence labelling datasets like SST-2 and AG News when compared to more complex tasks like RTE and MRPC which require comparison between sentences. Existing work \cite{pande2021heads} shows that for simpler tasks, the BERT heads perform discrete non-overlapping roles, while for complex tasks, there is greater overlap in head roles and a few heads perform more than one role. 
We hypothesize that this nature implies that the attention masks for different inputs even belonging to the same type (authentic or adversarial) can vary widely. This reduces the consistency of features across input types making the detection harder.
Nevertheless, AdvNet establishes state-of-the-art results across datasets. 
A detailed analysis of the performance of AdvNet across tasks and attack types is provided in Appendix \ref{appendix:results_per_attack_type}.

\subsection{Ablation studies}
\label{sec:ablation_studies}
We now evaluate how variations in model size, training set size, and the choice of feature combinations effect performance of AdvNet.


\noindent\textbf{Effect of model size.} 
IAS can be computed for Transformer networks of any size.
We compare BERT-Small and BERT-Base models in terms of performance of AdvNet as shown in Table \ref{tab:adnet_compare}. 
We observe that, across datasets, AdvNet performs better in detecting adversarial inputs fed to the larger BERT-Base model (108M parameters) as opposed to the smaller BERT-Small model (25M parameters). 
The increase in accuracy averaged across tasks is a significant 10.76\%.
We hypothesize that this is because models with more layers encode more information and allow for a better build-up of semantic information which means that individual heads play more discrete roles. 
This better performance for the larger model is encouraging as the more accurate and larger language models are expected to be more vulnerable to adversarial attacks.

\begin{figure}
    \centering
    \includegraphics[scale=0.45]{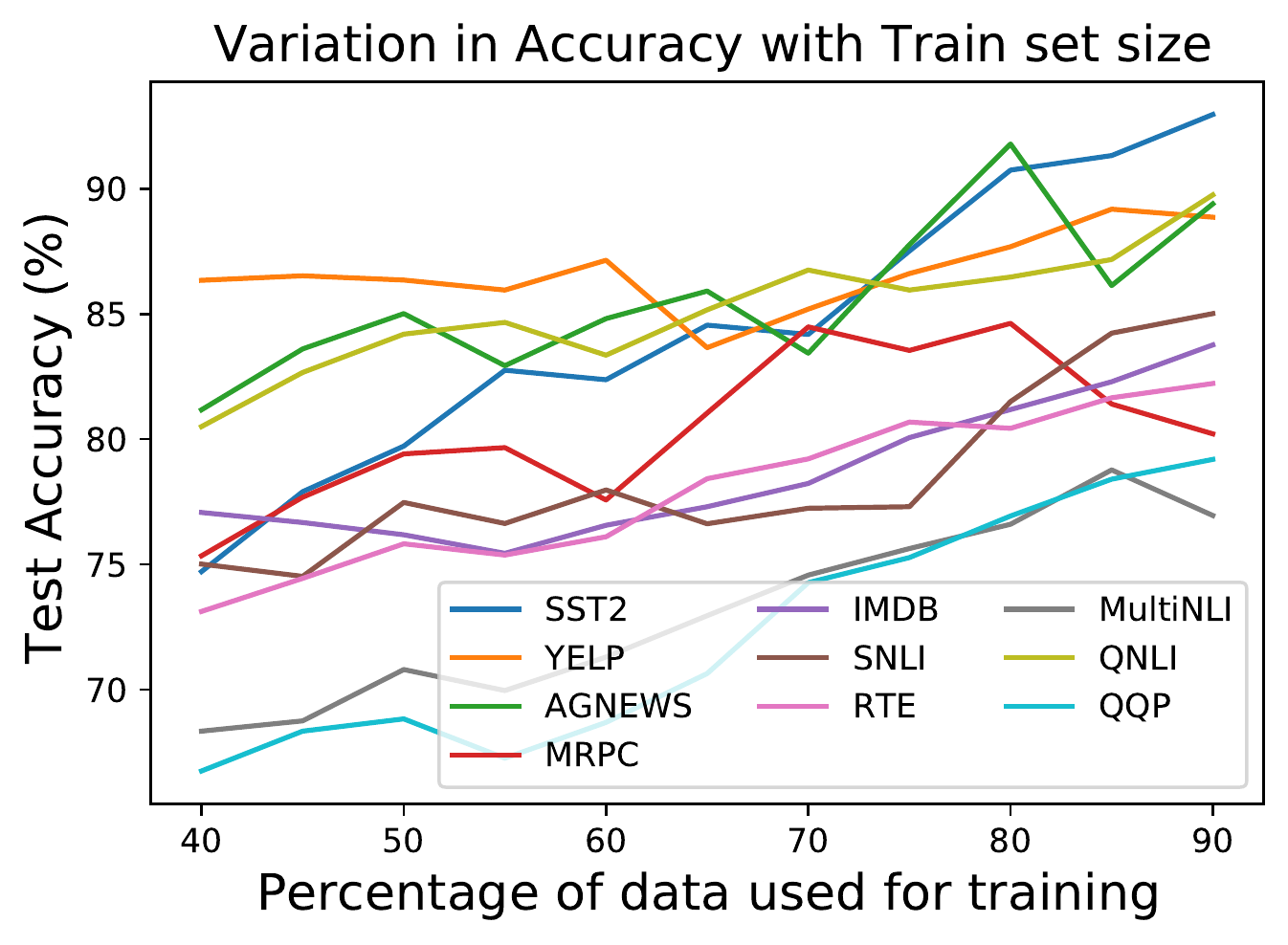}
    \caption{Effect of training set size on accuracy of adversarial detection with AdvNet.}
    \label{fig:effect_train_size}
\end{figure}

\noindent\textbf{Effect of training set size.} In Figure \ref{fig:effect_train_size}, we show how the performance of AdvNet changes as the amount of training data changes. We observe that AdvNet performs well even when it uses only a fraction of the training set. 
Specifically, even at 40\% of the training examples used, AdvNet out-performs the results obtained with existing state-of-the-art models on most tasks. 
This suggests that the CutMix data augmentation is effective and the AdvNet model is sample-efficient. 
This is particularly important because designing adversarial examples for each dataset remains a challenging task. 
\setlength{\tabcolsep}{3pt}
\begin{table}[!h]
\centering
\begin{tabular}{c c c c c c }
\hline
\textbf{Datasets}&\textbf{\fmask}&\textbf{\fflip}&\textbf{\flayer} &\textbf{Bin} &\textbf{w/o CM}\\\hline
{SST-2} & 82.87 & 74.07 & 64.79 & 85.59 & 82.23  \\ 
{Yelp} & 80.23 & 62.08 & 66.01 & 84.30 & 83.57  \\ 
{AG News} & 83.11 & 76.41 & 57.14 & 90.47 & 83.11 \\ 
{MRPC} & 76.35 & 68.82 & 59.40 & 80.27& 77.35  \\ 
{IMDb} & 74.54 & 60.00 & 55.45 & 73.78 & 74.23 \\ 
{SNLI} & 80.83 & 57.91 & 58.83 & 75.64& 70.41  \\ 
{RTE} & 74.44 & 60.88 & 56.67 & 77.21& 74.06  \\  
{MNLI} & 66.95 & 51.30 & 60.00 & 66.85& 69.95  \\ 
{QQP} & 66.41 & 61.63 & 62.64 & 71.88 & 64.50  \\ 
{QNLI} & 79.65 & 55.69 & 59.36 & 81.42 & 73.11 \\ \hline
\end{tabular}
\caption{Results on feature combinations.}
\label{tab:ablation_study}
\end{table}
\addtolength{\tabcolsep}{-2pt} 

\noindent\textbf{Using different feature combinations.}
We had shown that each of the three features are informative in Section \ref{sec:feat_sp_analysis}.
In Table \ref{tab:ablation_study}, we report the performance of AdvNet by ablating various model components. The first 3 columns report accuracies when only one of the three features is passed at a time to the model. We observe that \fmask~performs better than \fflip~and \flayer. This suggests that the attention mask is the most important feature input to the model. We analyze the roles of individual gating values using GradCAM (see Appendix \ref{sec:refereeing_heads}).
Next, we test the performance when the boolean attention mask \fbmask~is used instead of the real-valued vector \fmask~along with \fflip~and \flayer. The lower accuracy indicates that the real values are more informative. Finally, we test the model performance when CutMix is not used and conclude that augmenting the training set using CutMix provides higher accuracy as seen in the last row of Table \ref{tab:adnet_compare} which uses all 3 features along with CutMix.

\noindent\textbf{Defense Transferability Analysis.}
Next, we perform a study to understand how well the model can perform on unseen attack types. For this purpose, we train AdvNet with samples from only $x\%$ of the 11 attack types and report results both on test samples from the remaining attack types and the complete test set for $x \in \{25, 50, 75\}$ in Table \ref{tab:def_trans}. We observe that even when AdvNet is trained with only 75\% of the attack types, the test results on new attacks outperform existing approaches for most datasets, thus showing that our model can generalize to unseen attack methods. Besides, at all three values of $x$, the results on the complete test set closely agree with the results on the new attack types. This indicates that the reduction in accuracy at lower $x$ values can largely be attributed to a smaller training set than to a lack of defense transferability.

\setlength{\tabcolsep}{3pt}
\begin{table}[!h]
  \centering
 \begin{tabular}{c c c c c c c}
    \hline
 \textbf{\Longstack{Dataset}} & 
 \textbf{\Longstack{25\%}} & \textbf{\Longstack{50\%}} & \textbf{\Longstack{75\%}} \\ \hline
    SST-2 & (57.8, 58.9) & (69.9, 68.4) & (82.7, 80.7) \\
    Yelp & (63.1, 61.8)& (70.3, 69.9)& (77.8, 78.4) \\
    AG News &(63.7, 62.1) & (71.4, 69.9) & (83.6, 78.4)\\
    MRPC & (63.2, 60.1) & (73.2, 74.5) & (81.5, 82.3)\\ 
    IMDb & (66.8, 64.8) & (71.6, 73.1) & (77.4, 79.1)\\
    SNLI & (57.9, 57.6) &(67.2, 66.6) &(73.4, 72.3)  \\
    RTE & (63.8, 62.4) &(70.8, 69.7)  &(76.4, 75.5) \\
    MNLI &(57.4, 58.8) &(62.3, 61.3) &(67.0, 68.9) \\
    QQP & (59.7, 60.2) &(64.0, 64.2) & (69.0, 69.6) \\
    QNLI & (61.8, 60.6) & (69.3, 67.2) & (75.7, 77.5) \\
    \hline
  \end{tabular}
  \caption{Defense transferability study of AdvNet with varying percentages of attack types included in the train set. Each tuple contains the test accuracy on new attack types and on all attack types respectively.} \label{tab:def_trans}
\end{table}

In summary, our results show that (a) the 3 IAS features are individually informative, (b) AdvNet significantly improves on baseline methods across  datasets, (c) AdvNet performance improves with model size and does not drop much on reducing training sets, (d) AdvNet achieves the best performance when all 3 features are used along with CutMix augmentation, and (e) AdvNet generalizes well to new attack types. 

\section{Conclusion and future work}\label{sec:conclusion}
In this work, we present an altogether new utility of attention heads in Transformer networks - to detect adversarial attacks. 
We defined input-specific attention subnetworks (IAS) and proposed a method to compute them efficiently. 
We extracted 3 features from IAS and showed their utility in distinguishing adversarial samples from authentic ones. 
We demonstrated that our approach significantly improves the state-of-the-art accuracy across datasets and attack types.
Our work suggests that input-specific model perturbations provide strong signals to interpret Transformer-based models such as large language models.
Further, the sparse nature of the identified IAS indicate opportunities for input-specific model optimization.
In future work, we would like to extend this study to tasks beyond NLU, including vision and speech-related tasks.

\section*{Discussion on Ethics and broader impact}
One of the main challenges with deep neural models is their lack of explainability. These models typically have inherent biases resulting from the training data, parameter combinations and other factors that lead to unexpected responses to certain inputs. This is further complicated when adversarial agents target to manipulate the output of deep neural models. We see our work on creating and using attention subnetworks for adversarial detection as a part of the broader effort towards Responsible AI. Such a solution is particularly important in situations where deep neural models make decisions that affect physical safety, digital security and equal opportunity. However, we acknowledge that this additional visibility into the model comes at an added cost - inference under uncertainty of adversarial detection is more expensive. We encourage system designers to trade-off computational and runtime considerations for security when deploying such solutions.


\section*{Acknowledgements}
We thank Samsung and IITM Pravartak for supporting our work through their joint fellowship program. We also wish to thank the anonymous reviewers for their efforts in evaluating our work and providing us with constructive feedback.


\bibliography{anthology,custom}

\begin{thebibliography}{39}
\expandafter\ifx\csname natexlab\endcsname\relax\def\natexlab#1{#1}\fi

\bibitem[{Biju et~al.(2020)Biju, Sriram, Khapra, and
  Kumar}]{biju-etal-2020-joint}
Emil Biju, Anirudh Sriram, Mitesh~M. Khapra, and Pratyush Kumar. 2020.
\newblock \href {https://doi.org/10.18653/v1/2020.coling-main.87} {Joint
  transformer/{RNN} architecture for gesture typing in indic languages}.
\newblock In \emph{Proceedings of the 28th International Conference on
  Computational Linguistics}, pages 999--1010, Barcelona, Spain (Online).
  International Committee on Computational Linguistics.

\bibitem[{Bowman et~al.(2015)Bowman, Angeli, Potts, and
  Manning}]{bowman-etal-2015-large}
Samuel~R. Bowman, Gabor Angeli, Christopher Potts, and Christopher~D. Manning.
  2015.
\newblock \href {https://doi.org/10.18653/v1/D15-1075} {A large annotated
  corpus for learning natural language inference}.
\newblock In \emph{Proceedings of the 2015 Conference on Empirical Methods in
  Natural Language Processing}, pages 632--642, Lisbon, Portugal. Association
  for Computational Linguistics.

\bibitem[{Budhraja et~al.(2020)Budhraja, Pande, Nema, Kumar, and
  Khapra}]{budhraja-etal-2020-weak}
Aakriti Budhraja, Madhura Pande, Preksha Nema, Pratyush Kumar, and Mitesh~M.
  Khapra. 2020.
\newblock \href {https://doi.org/10.18653/v1/2020.emnlp-main.260} {On the weak
  link between importance and prunability of attention heads}.
\newblock In \emph{Proceedings of the 2020 Conference on Empirical Methods in
  Natural Language Processing (EMNLP)}, pages 3230--3235, Online. Association
  for Computational Linguistics.

\bibitem[{Dolan and Brockett(2005)}]{dolan-brockett-2005-automatically}
William~B. Dolan and Chris Brockett. 2005.
\newblock \href {https://www.aclweb.org/anthology/I05-5002} {Automatically
  constructing a corpus of sentential paraphrases}.
\newblock In \emph{Proceedings of the Third International Workshop on
  Paraphrasing ({IWP}2005)}.

\bibitem[{Feng et~al.(2018)Feng, Wallace, Grissom~II, Iyyer, Rodriguez, and
  Boyd-Graber}]{feng-etal-2018-pathologies}
Shi Feng, Eric Wallace, Alvin Grissom~II, Mohit Iyyer, Pedro Rodriguez, and
  Jordan Boyd-Graber. 2018.
\newblock \href {https://doi.org/10.18653/v1/D18-1407} {Pathologies of neural
  models make interpretations difficult}.
\newblock In \emph{Proceedings of the 2018 Conference on Empirical Methods in
  Natural Language Processing}, pages 3719--3728, Brussels, Belgium.
  Association for Computational Linguistics.

\bibitem[{Gao et~al.(2018)Gao, Lanchantin, Soffa, and Qi}]{gao2018black}
Ji~Gao, Jack Lanchantin, Mary~Lou Soffa, and Yanjun Qi. 2018.
\newblock Black-box generation of adversarial text sequences to evade deep
  learning classifiers.
\newblock In \emph{2018 IEEE Security and Privacy Workshops (SPW)}, pages
  50--56. IEEE.

\bibitem[{Goldberg(2019)}]{goldberg2019assessing}
Yoav Goldberg. 2019.
\newblock Assessing bert's syntactic abilities.
\newblock \emph{arXiv preprint arXiv:1901.05287}.

\bibitem[{Hewitt and Manning(2019)}]{hewitt-manning-2019-structural}
John Hewitt and Christopher~D. Manning. 2019.
\newblock \href {https://doi.org/10.18653/v1/N19-1419} {{A} structural probe
  for finding syntax in word representations}.
\newblock In \emph{Proceedings of the 2019 Conference of the North {A}merican
  Chapter of the Association for Computational Linguistics: Human Language
  Technologies, Volume 1 (Long and Short Papers)}, pages 4129--4138,
  Minneapolis, Minnesota. Association for Computational Linguistics.

\bibitem[{Jawahar et~al.(2019)Jawahar, Sagot, and Seddah}]{jawahar2019does}
Ganesh Jawahar, Beno{\^\i}t Sagot, and Djam{\'e} Seddah. 2019.
\newblock What does bert learn about the structure of language?
\newblock In \emph{ACL 2019-57th Annual Meeting of the Association for
  Computational Linguistics}.

\bibitem[{Jia et~al.(2019)Jia, Raghunathan, G{\"o}ksel, and
  Liang}]{Jia2019CertifiedRT}
Robin Jia, Aditi Raghunathan, Kerem G{\"o}ksel, and Percy Liang. 2019.
\newblock Certified robustness to adversarial word substitutions.
\newblock In \emph{EMNLP}.

\bibitem[{Jiao et~al.(2020)Jiao, Yin, Shang, Jiang, Chen, Li, Wang, and
  Liu}]{jiao-etal-2020-tinybert}
Xiaoqi Jiao, Yichun Yin, Lifeng Shang, Xin Jiang, Xiao Chen, Linlin Li, Fang
  Wang, and Qun Liu. 2020.
\newblock \href {https://doi.org/10.18653/v1/2020.findings-emnlp.372}
  {{T}iny{BERT}: Distilling {BERT} for natural language understanding}.
\newblock In \emph{Findings of the Association for Computational Linguistics:
  EMNLP 2020}, pages 4163--4174, Online. Association for Computational
  Linguistics.

\bibitem[{Jin et~al.(2020)Jin, Jin, Zhou, and
  Szolovits}]{DBLP:conf/aaai/JinJZS20}
Di~Jin, Zhijing Jin, Joey~Tianyi Zhou, and Peter Szolovits. 2020.
\newblock \href {https://aaai.org/ojs/index.php/AAAI/article/view/6311} {Is
  {BERT} really robust? {A} strong baseline for natural language attack on text
  classification and entailment}.
\newblock In \emph{The Thirty-Fourth {AAAI} Conference on Artificial
  Intelligence, {AAAI} 2020, The Thirty-Second Innovative Applications of
  Artificial Intelligence Conference, {IAAI} 2020, The Tenth {AAAI} Symposium
  on Educational Advances in Artificial Intelligence, {EAAI} 2020, New York,
  NY, USA, February 7-12, 2020}, pages 8018--8025. {AAAI} Press.

\bibitem[{Louizos et~al.(2017)Louizos, Welling, and
  Kingma}]{louizos2017learning}
Christos Louizos, Max Welling, and Diederik~P Kingma. 2017.
\newblock Learning sparse neural networks through $ l\_0 $ regularization.
\newblock \emph{arXiv preprint arXiv:1712.01312}.

\bibitem[{Maas et~al.(2011)Maas, Daly, Pham, Huang, Ng, and
  Potts}]{maas-EtAl:2011:ACL-HLT2011}
Andrew~L. Maas, Raymond~E. Daly, Peter~T. Pham, Dan Huang, Andrew~Y. Ng, and
  Christopher Potts. 2011.
\newblock \href {http://www.aclweb.org/anthology/P11-1015} {Learning word
  vectors for sentiment analysis}.
\newblock In \emph{Proceedings of the 49th Annual Meeting of the Association
  for Computational Linguistics: Human Language Technologies}, pages 142--150,
  Portland, Oregon, USA. Association for Computational Linguistics.

\bibitem[{Michel et~al.(2019)Michel, Levy, and Neubig}]{NEURIPS2019_2c601ad9}
Paul Michel, Omer Levy, and Graham Neubig. 2019.
\newblock \href
  {https://proceedings.neurips.cc/paper/2019/file/2c601ad9d2ff9bc8b282670cdd54f69f-Paper.pdf}
  {Are sixteen heads really better than one?}
\newblock In \emph{Advances in Neural Information Processing Systems},
  volume~32. Curran Associates, Inc.

\bibitem[{Morris et~al.(2020)Morris, Lifland, Yoo, Grigsby, Jin, and
  Qi}]{morris2020textattack}
John~X. Morris, Eli Lifland, Jin~Yong Yoo, Jake Grigsby, Di~Jin, and Yanjun Qi.
  2020.
\newblock \href {http://arxiv.org/abs/2005.05909} {Textattack: A framework for
  adversarial attacks, data augmentation, and adversarial training in nlp}.

\bibitem[{Mozes et~al.(2021)Mozes, Stenetorp, Kleinberg, and
  Griffin}]{mozes-etal-2021-frequency}
Maximilian Mozes, Pontus Stenetorp, Bennett Kleinberg, and Lewis Griffin. 2021.
\newblock \href {https://www.aclweb.org/anthology/2021.eacl-main.13}
  {Frequency-guided word substitutions for detecting textual adversarial
  examples}.
\newblock In \emph{Proceedings of the 16th Conference of the European Chapter
  of the Association for Computational Linguistics: Main Volume}, pages
  171--186, Online. Association for Computational Linguistics.

\bibitem[{Mrk{\v{s}}i{\'c} et~al.(2016)Mrk{\v{s}}i{\'c}, S{\'e}aghdha, Thomson,
  Ga{\v{s}}i{\'c}, Rojas-Barahona, Su, Vandyke, Wen, and
  Young}]{mrkvsic2016counter}
Nikola Mrk{\v{s}}i{\'c}, Diarmuid~O S{\'e}aghdha, Blaise Thomson, Milica
  Ga{\v{s}}i{\'c}, Lina Rojas-Barahona, Pei-Hao Su, David Vandyke, Tsung-Hsien
  Wen, and Steve Young. 2016.
\newblock Counter-fitting word vectors to linguistic constraints.
\newblock \emph{arXiv preprint arXiv:1603.00892}.

\bibitem[{M\"{u}ller et~al.(2019)M\"{u}ller, Kornblith, and
  Hinton}]{NEURIPS2019_f1748d6b}
Rafael M\"{u}ller, Simon Kornblith, and Geoffrey~E Hinton. 2019.
\newblock \href
  {https://proceedings.neurips.cc/paper/2019/file/f1748d6b0fd9d439f71450117eba2725-Paper.pdf}
  {When does label smoothing help?}
\newblock In \emph{Advances in Neural Information Processing Systems},
  volume~32. Curran Associates, Inc.

\bibitem[{Pande et~al.(2021)Pande, Budhraja, Nema, Kumar, and
  Khapra}]{pande2021heads}
Madhura Pande, Aakriti Budhraja, Preksha Nema, Pratyush Kumar, and Mitesh~M
  Khapra. 2021.
\newblock The heads hypothesis: A unifying statistical approach towards
  understanding multi-headed attention in bert.
\newblock \emph{arXiv preprint arXiv:2101.09115}.

\bibitem[{Pruthi et~al.(2019)Pruthi, Dhingra, and Lipton}]{pruthi2019combating}
Danish Pruthi, Bhuwan Dhingra, and Zachary~C Lipton. 2019.
\newblock Combating adversarial misspellings with robust word recognition.
\newblock \emph{arXiv preprint arXiv:1905.11268}.

\bibitem[{Rajpurkar et~al.(2016)Rajpurkar, Zhang, Lopyrev, and
  Liang}]{rajpurkar-etal-2016-squad}
Pranav Rajpurkar, Jian Zhang, Konstantin Lopyrev, and Percy Liang. 2016.
\newblock \href {https://doi.org/10.18653/v1/D16-1264} {{SQ}u{AD}: 100,000+
  questions for machine comprehension of text}.
\newblock In \emph{Proceedings of the 2016 Conference on Empirical Methods in
  Natural Language Processing}, pages 2383--2392, Austin, Texas. Association
  for Computational Linguistics.

\bibitem[{Ren et~al.(2019)Ren, Deng, He, and Che}]{ren-etal-2019-generating}
Shuhuai Ren, Yihe Deng, Kun He, and Wanxiang Che. 2019.
\newblock \href {https://doi.org/10.18653/v1/P19-1103} {Generating natural
  language adversarial examples through probability weighted word saliency}.
\newblock In \emph{Proceedings of the 57th Annual Meeting of the Association
  for Computational Linguistics}, pages 1085--1097, Florence, Italy.
  Association for Computational Linguistics.

\bibitem[{{Selvaraju} et~al.(2017){Selvaraju}, {Cogswell}, {Das}, {Vedantam},
  {Parikh}, and {Batra}}]{8237336}
R.~R. {Selvaraju}, M.~{Cogswell}, A.~{Das}, R.~{Vedantam}, D.~{Parikh}, and
  D.~{Batra}. 2017.
\newblock \href {https://doi.org/10.1109/ICCV.2017.74} {Grad-cam: Visual
  explanations from deep networks via gradient-based localization}.
\newblock In \emph{2017 IEEE International Conference on Computer Vision
  (ICCV)}, pages 618--626.

\bibitem[{Socher et~al.(2013)Socher, Perelygin, Wu, Chuang, Manning, Ng, and
  Potts}]{socher-etal-2013-recursive}
Richard Socher, Alex Perelygin, Jean Wu, Jason Chuang, Christopher~D. Manning,
  Andrew Ng, and Christopher Potts. 2013.
\newblock \href {https://www.aclweb.org/anthology/D13-1170} {Recursive deep
  models for semantic compositionality over a sentiment treebank}.
\newblock In \emph{Proceedings of the 2013 Conference on Empirical Methods in
  Natural Language Processing}, pages 1631--1642, Seattle, Washington, USA.
  Association for Computational Linguistics.

\bibitem[{Tenney et~al.(2019)Tenney, Das, and Pavlick}]{tenney-etal-2019-bert}
Ian Tenney, Dipanjan Das, and Ellie Pavlick. 2019.
\newblock \href {https://doi.org/10.18653/v1/P19-1452} {{BERT} rediscovers the
  classical {NLP} pipeline}.
\newblock In \emph{Proceedings of the 57th Annual Meeting of the Association
  for Computational Linguistics}, pages 4593--4601, Florence, Italy.
  Association for Computational Linguistics.

\bibitem[{van Aken et~al.(2019)van Aken, Winter, L\"{o}ser, and
  Gers}]{10.1145/3357384.3358028}
Betty van Aken, Benjamin Winter, Alexander L\"{o}ser, and Felix~A. Gers. 2019.
\newblock \href {https://doi.org/10.1145/3357384.3358028} {How does bert answer
  questions? a layer-wise analysis of transformer representations}.
\newblock CIKM '19, page 1823–1832, New York, NY, USA. Association for
  Computing Machinery.

\bibitem[{van~der Maaten and Hinton(2008)}]{JMLR:v9:vandermaaten08a}
Laurens van~der Maaten and Geoffrey Hinton. 2008.
\newblock \href {http://jmlr.org/papers/v9/vandermaaten08a.html} {Visualizing
  data using t-sne}.
\newblock \emph{Journal of Machine Learning Research}, 9(86):2579--2605.

\bibitem[{Voita et~al.(2019)Voita, Talbot, Moiseev, Sennrich, and
  Titov}]{voita2019analyzing}
Elena Voita, David Talbot, Fedor Moiseev, Rico Sennrich, and Ivan Titov. 2019.
\newblock Analyzing multi-head self-attention: Specialized heads do the heavy
  lifting, the rest can be pruned.
\newblock \emph{arXiv preprint arXiv:1905.09418}.

\bibitem[{Wang et~al.(2018)Wang, Singh, Michael, Hill, Levy, and
  Bowman}]{wang-etal-2018-glue}
Alex Wang, Amanpreet Singh, Julian Michael, Felix Hill, Omer Levy, and Samuel
  Bowman. 2018.
\newblock \href {https://doi.org/10.18653/v1/W18-5446} {{GLUE}: A multi-task
  benchmark and analysis platform for natural language understanding}.
\newblock In \emph{Proceedings of the 2018 {EMNLP} Workshop {B}lackbox{NLP}:
  Analyzing and Interpreting Neural Networks for {NLP}}, pages 353--355,
  Brussels, Belgium. Association for Computational Linguistics.

\bibitem[{Wang et~al.(2019)Wang, Dong, Sun, Wang, and
  Zhang}]{wang2019adversarial}
Jingyi Wang, Guoliang Dong, Jun Sun, Xinyu Wang, and Peixin Zhang. 2019.
\newblock Adversarial sample detection for deep neural network through model
  mutation testing.
\newblock In \emph{2019 IEEE/ACM 41st International Conference on Software
  Engineering (ICSE)}, pages 1245--1256. IEEE.

\bibitem[{Wang et~al.(2020)Wang, Su, Zhang, and Hu}]{wang2020interpret}
Yulong Wang, Hang Su, Bo~Zhang, and Xiaolin Hu. 2020.
\newblock Interpret neural networks by extracting critical subnetworks.
\newblock \emph{IEEE Transactions on Image Processing}, 29:6707--6720.

\bibitem[{Williams et~al.(2018)Williams, Nangia, and
  Bowman}]{williams-etal-2018-broad}
Adina Williams, Nikita Nangia, and Samuel Bowman. 2018.
\newblock \href {https://doi.org/10.18653/v1/N18-1101} {A broad-coverage
  challenge corpus for sentence understanding through inference}.
\newblock In \emph{Proceedings of the 2018 Conference of the North {A}merican
  Chapter of the Association for Computational Linguistics: Human Language
  Technologies, Volume 1 (Long Papers)}, pages 1112--1122, New Orleans,
  Louisiana. Association for Computational Linguistics.

\bibitem[{Xie et~al.(2019)Xie, Wu, Maaten, Yuille, and He}]{xie2019feature}
Cihang Xie, Yuxin Wu, Laurens van~der Maaten, Alan~L Yuille, and Kaiming He.
  2019.
\newblock Feature denoising for improving adversarial robustness.
\newblock In \emph{Proceedings of the IEEE/CVF Conference on Computer Vision
  and Pattern Recognition}, pages 501--509.

\bibitem[{Yun et~al.(2019)Yun, Han, Oh, Chun, Choe, and Yoo}]{yun2019cutmix}
Sangdoo Yun, Dongyoon Han, Seong~Joon Oh, Sanghyuk Chun, Junsuk Choe, and
  Youngjoon Yoo. 2019.
\newblock Cutmix: Regularization strategy to train strong classifiers with
  localizable features.
\newblock In \emph{Proceedings of the IEEE/CVF International Conference on
  Computer Vision}, pages 6023--6032.

\bibitem[{Zhang et~al.(2015{\natexlab{a}})Zhang, Zhao, and
  LeCun}]{zhangCharacterlevelConvolutionalNetworks2015}
Xiang Zhang, Junbo Zhao, and Yann LeCun. 2015{\natexlab{a}}.
\newblock \href {http://arxiv.org/abs/1509.01626} {Character-level
  {{Convolutional Networks}} for {{Text Classification}}}.
\newblock \emph{arXiv:1509.01626 [cs]}.

\bibitem[{Zhang et~al.(2015{\natexlab{b}})Zhang, Zhao, and
  LeCun}]{Zhang2015CharacterlevelCN}
Xiang Zhang, Junbo~Jake Zhao, and Yann LeCun. 2015{\natexlab{b}}.
\newblock Character-level convolutional networks for text classification.
\newblock In \emph{NIPS}.

\bibitem[{Zhou et~al.(2019)Zhou, Jiang, Chang, and
  Wang}]{zhou-etal-2019-learning}
Yichao Zhou, Jyun-Yu Jiang, Kai-Wei Chang, and Wei Wang. 2019.
\newblock \href {https://doi.org/10.18653/v1/D19-1496} {Learning to
  discriminate perturbations for blocking adversarial attacks in text
  classification}.
\newblock In \emph{Proceedings of the 2019 Conference on Empirical Methods in
  Natural Language Processing and the 9th International Joint Conference on
  Natural Language Processing (EMNLP-IJCNLP)}, pages 4904--4913, Hong Kong,
  China. Association for Computational Linguistics.

\bibitem[{Zhu et~al.(2019)Zhu, Cheng, Gan, Sun, Goldstein, and
  Liu}]{Zhu2019FreeLBEA}
Chen Zhu, Yu~Cheng, Zhe Gan, S.~Sun, Tom Goldstein, and Jingjing Liu. 2019.
\newblock Freelb: Enhanced adversarial training for language understanding.
\newblock \emph{ArXiv}, abs/1909.11764.

\end{thebibliography}
\bibliographystyle{acl_natbib}
\newpage
\appendix
\section{Datasets used for authentic examples}
\label{appendix:datasets}
The 10 datasets used in this work were listed in Section \ref{sec:data_sec}. Here, we provide additional details about these datasets. SST-2 \cite{socher-etal-2013-recursive},
Yelp polarity \cite{zhangCharacterlevelConvolutionalNetworks2015} and IMDb \cite{maas-EtAl:2011:ACL-HLT2011} are binary sentiment classification datasets. AG News \cite{Zhang2015CharacterlevelCN} consists of news headlines classified into one of 4 categories (world, sports, business, sci/tech) and MRPC \cite{dolan-brockett-2005-automatically} is a paraphrase dataset which contains sentence pairs with binary labels indicating whether they are semantically equivalent or not.
RTE \cite{wang-etal-2018-glue}, 
MNLI \cite{williams-etal-2018-broad}, SNLI \cite{bowman-etal-2015-large} contain sentence pairs with labels indicating whether one sentences entails, contradicts or is neutral with respect to the other sentence. QQP is again a paraphrase dataset but unlike MRPC which contains sentences, it contains question pairs taken from Quora with binary labels indicating whether they are semantically equivalent or not. QNLI contains question-context pairs with a binary label indicating whether the context sentence contains the answer to the question or not.

\section{Examples of adversarial attacks} \label{sec:examples_adv_attacks}
In Table \ref{tab:adv_samples}, we provide examples for each of the 11 attack types that we use to generate adversarial inputs for this work. 

\begin{table*}[!h]
    \centering
    \begin{tabular}{|c|c|}\hline
    \textbf{\Longstack{Attack Type}} &  \textbf{\Longstack{Perturbed Text}} \\ \hline
    \textbf{\Longstack{Original Text}} &  \textbf{\Longstack{it 's a charming and often affecting journey.}} \\ \hline 
        \textbf{Word-level attacks} & \\
        Deletion & it's a \MK{\_} and often affecting journey. \\
        Antonyms & it's a \MK{repulsive} and often affecting journey. \\
        Synonyms & it's a charming and often affecting \MK{passage}. \\
        Embeddings & it's a charming and \MK{quite} affecting journey. \\
        Order Swap & it's charming and \MK{affecting} \MK{a} \MK{often} journey. \\
        PWWS & it's a \MK{entrance} and often \MK{strike} journey. \\
        TextFooler & it's a charming and \MK{\_} affecting journey. \\ \hline
        \textbf{\Longstack{Original Text}} &  \textbf{\Longstack{a sometimes tedious film.}} \\ \hline 
        \textbf{Character-level attacks} & \\
        Substitution & a sometimes t\MK{i}dious f\MK{y}lm.\\
        Deletion & a som\MK{\_}times tedio\MK{\_}s film. \\
        Insertion & a sometime\MK{D}s t\MK{v}edious film.\\
        Order Swap & a s\MK{mo}etimes ted\MK{oi}us film. \\ \hline
    \end{tabular}
    \caption{Examples of 11 attack types used for adversarial data creation. `\MK{\_}' represents a deleted character and there is no character present at that position in the adversarial sample.}
    \label{tab:adv_samples}
\end{table*}

\section{Other methods for Adversarial Detection}
\label{appendix:sota_methods}
We briefly describe the four methods that we compare with in Table \ref{tab:adnet_compare}.

\begin{itemize}[noitemsep, leftmargin=0.4cm, topsep=1pt]
\item \textbf{FGWS} \citep{mozes-etal-2021-frequency}: Here, a word frequency-guided approach is used to identify infrequent words in an input sentence and replace them with more frequent, semantically similar words. Then, the difference in prediction confidence of the Transformer-based model between the original and substituted sentences is considered. If this value is above a threshold, the sentence is predicted to be adversarial.
\item \textbf{NWS}: This is the \textit{naive word substitution} baseline used in \citet{mozes-etal-2021-frequency}. Here, each out-of-vocabulary word in an input sentence is replaced with a random word from a set of semantically related words, following which the same process as above is used to predict input authenticity. 
\item \textbf{DISP} \citep{zhou-etal-2019-learning}: In this approach, a BERT-based perturbation discriminator predicts whether each token in the input sentence is authentic or perturbed. If none of the tokens are predicted to be perturbed, the input sentence is considered authentic.
\item \textbf{FreeLB}
\citep{Zhu2019FreeLBEA}: This is an adversarial training approach where adversarial perturbations are added to word embeddings and the resulting adversarial loss is minimized to promote higher invariance in the embedding space.

\item \textbf{Certified Robustness Training}
\citep{Jia2019CertifiedRT}: 
This approach uses Interval Bound Propagation (IBP) to obtain an upper bound on the worst-case loss resulting from any word substitution-based perturbation. This has been applied to CNN and LSTM-based language models.
\end{itemize}

\begin{table}
    \centering
    \begin{tabular}{c c c}
    \hline
    \textbf{Dataset} & \textbf{\Longstack{Non-target o/p\\(Mutated)}} & \textbf{\Longstack{Switches$>$1\\(Layer-wise)}}\\ \hline
         SST-2 &  (12.3, 34.2) & (37.9, 54.8)\\
         IMDb &  (0.33, 2.18) & (0.16, 1.45)\\
         Yelp & (3.8, 5.3) & (0.83, 1.08)\\
         AG News & (6.6, 22.8) & (3.2, 17.0)\\
         MRPC & (21.3, 24.3) & (10.3, 8.77)\\
         RTE & (24.5, 22.2) & (44.2, 50.9)\\
         SNLI & (2.83, 96.0)  & (11.6, 41.0) \\
         MNLI & (11.0, 24.8) & (24.3, 42.5) \\
         QQP & (3.2, 1.3) & (6.2, 6.8) \\
         QNLI & (5.7, 1.0) & (13.8, 11.1) \\\hline
    \end{tabular}
    \caption{Percentages of (authentic, adversarial) inputs whose (a) mutated subnetworks generated non-target class predictions; (b) layer-wise outputs showed more than one switch.}
    \label{tab:feat23}
\end{table}

\section{Analysing \fflip~and \flayer}
\label{appendix:analysing_fflip}
In the second column of Table \ref{tab:feat23}, for each of the datasets, we show the percentage of authentic and adversarial inputs which generated non-target class predictions. Further, in the third column of Table \ref{tab:feat23} we show the percentage of (authentic, adversarial) inputs whose layer-wise outputs showed more than one switch. 
These results show that the \fflip~and \flayer~are individually informative.

\section{Adversarial detection accuracy for different attack types}
\label{appendix:results_per_attack_type}
In Table \ref{tab:attack_acc}, we present the breakup of model accuracy across individual attack types. We observe that for text classification tasks like SST-2, Yelp and AG News the accuracy for Embedding and Synonym swap attack types are much higher compared to other datasets. We also note that in case of both word and character-level attacks, Deletion and Substitution operations are the ones with least detection accuracy across almost all datasets. Finally, we observe that the performance for detecting adversarial inputs generated by PWWS and TextFooler attacks remain fairly consistent across datasets.
\makeatletter
    \setlength\@fptop{0\p@}
\makeatother
\begin{table*}[!t]
  \centering
 \begin{tabular}{c | c | c c c c c c c | c c c c}
    \hline
 \textbf{\Longstack{Dataset}}&\textbf{\Longstack{\#Adv}}& \multicolumn{7}{c}{\textbf{\Longstack{Word-level attacks}}} & \multicolumn{4}{c}{\textbf{\Longstack{Character-level attacks}}} \\ \cline{3-13}
    & \textbf{samples}
    & \textbf{\Longstack{DEL}} & \textbf{\Longstack{ANT}} & \textbf{\Longstack{SYN}} &\textbf{\Longstack{EMBED}} & \textbf{\Longstack{SWAP}} & \textbf{\Longstack{PWWS}} & \textbf{\Longstack{TF}} & \textbf{\Longstack{SUB}} & \textbf{\Longstack{DEL}} & \textbf{\Longstack{INS}} & \textbf{\Longstack{SWAP}}\\
    \hline
    SST-2 & 739 & 0.84 & 0.96 & 0.95 & 0.96 & 0.75 & 0.81 & 0.76 & 0.92 & 0.80 & 0.87 & 0.89 \\
    Yelp & 589 & 0.75 & 0.92 & 0.92 & 0.96 & 0.88  & 0.80 & 0.95 & 0.93 & 0.77 & 0.88 & 0.88 \\
    AG News & 829 & 0.88 & 0.96 & 0.92 & 0.96 & 0.82 & 0.83 & 0.84 & 0.89 & 0.84 & 0.85 & 0.88 \\
    MRPC & 712 & 0.75 & 0.75 & 0.9 & 0.72 & 0.94 & 0.84 & 0.82 & 0.86 & 0.79 & 0.76 & 0.92 \\ 
    IMDb & 321 & 0.80 & 0.76 & 0.85 & 0.89 & 0.80 & 0.82 & 0.81 & 0.94 & 0.75 & 0.96 & 0.79 \\
    SNLI & 1262 & 0.61 & 0.80 & 0.78 & 0.88 & 0.78 & 0.76 & 0.79 & 0.85 & 0.88 & 0.65 & 0.83 \\
    RTE & 541 & 0.75 & 0.84 & 0.86 & 0.87 & 0.79 & 0.77 & 0.73 & 0.82 & 0.76 & 0.82 & 0.82 \\
     MNLI & 548 & 0.67 & 0.80 & 0.72 & 0.85 & 0.78 & 0.80 & 0.76 & 0.78 & 0.80 & 0.86 & 0.76 \\
     QQP & 307 & 0.70 & 0.82 & 0.74 & 0.80 & 0.75 & 0.76 & 0.74 & 0.78 & 0.81 & 0.86 & 0.77 \\
     QNLI & 395 & 0.80 & 0.90 & 0.92 & 0.92 & 0.90 & 0.82 & 0.86 & 0.82 & 0.86 & 0.82 & 0.82 \\
     \hline
  \end{tabular}
  \caption{Accuracies across datasets for each attack type. \textbf{Legend}: SUB-substitution, DEL-deletion, SYN-synonym, EMBED-embedding, INS-insertion, SWAP-order swap, TF-TextFooler. Refer Section \ref{sec:data_sec} for descriptions of attack types. The second column provides the number of adversarial samples generated by us for each task across all 11 attack types.}\label{tab:attack_acc}
\end{table*}
\addtolength{\tabcolsep}{2pt} 

\section{Refereeing heads in adversarial detection} \label{sec:refereeing_heads}
\begin{figure}[!th]
    \centering
    \includegraphics[scale=0.52]{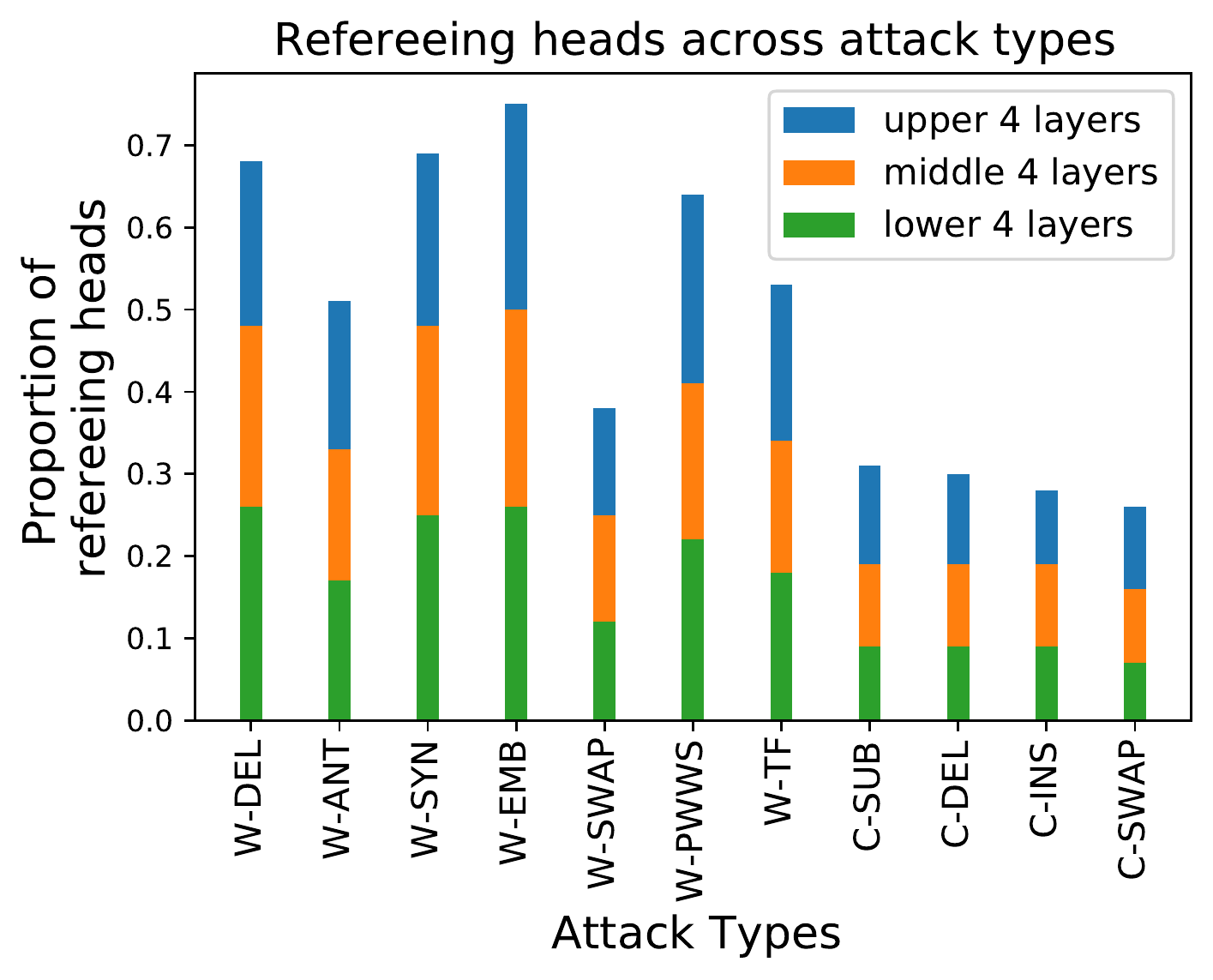}
    \caption{Fraction of refereeing heads used by the adversarial detection model across various adversarial attack types. The split of these across 4 layer subsets is also shown.}
    \label{fig:imp_and_refereeing_heads}
\end{figure}

In this section, we explore the influence of each gating value in generating the prediction for our adversarial detection model. We make use of the Grad-CAM \cite{8237336} approach to identify critical neurons in the input layer of AdvNet that have large gradients from the target class (authentic or adversarial) flowing through them. Among these, we consider neurons that correspond to the gating values, i.e, \fmask~and call the heads corresponding to them as \textit{refereeing} heads. 
From Figure \ref{fig:imp_and_refereeing_heads}, we observe that word swap attacks like antonyms, synonyms, and embeddings require a greater number of refereeing heads, while character-level attacks need fewer. This is because character-level changes make the token invalid, i.e, the model treats it as a unknown token absent in the vocabulary. Since this changes the input embedding sequence more dramatically \cite{biju-etal-2020-joint}, small deviations from standard gating patterns are sufficient to mislead the model leading to fewer refereeing heads. Since introducing synonym and embedding based perturbations change the embeddings input to the model by a smaller extent, larger deviations from the gating pattern are required to block or pass selective chunks of information to mislead the model.

\end{document}